\documentclass[12pt]{article}
\usepackage{graphicx}
\usepackage[utf8]{inputenc} % allow utf-8 input
\usepackage[T1]{fontenc}    % use 8-bit T1 fonts
\usepackage{hyperref}       % hyperlinks
\usepackage{url}            % simple URL typesetting
\usepackage{booktabs}       % professional-quality tables
\usepackage{amsfonts}       % blackboard math symbols
\usepackage{nicefrac}       % compact symbols for 1/2, etc.
\usepackage{microtype}      % microtypography
\usepackage{amsmath}
\usepackage{subcaption}
\usepackage{placeins}
\usepackage{float}
\floatstyle{ruled}
\newfloat{algorithm}{thp}{lop}
\floatname{algorithm}{Algorithm}

\usepackage{algorithmicx}
\usepackage{algpseudocode}

\usepackage{amsmath, amsfonts,amsthm,mathrsfs,amssymb}
\usepackage{enumerate}
\usepackage{natbib}
\usepackage{color}
\usepackage{url} % not crucial - just used below for the URL 
\usepackage{multirow}
\renewcommand{\baselinestretch}{1.3}

\usepackage[paperwidth=8.5in, paperheight=11in,
top=2.55cm, bottom=2.55cm, left=2.55cm, right=2.55cm]{geometry}
\begin{document}
\def\spacingset#1{\renewcommand{\baselinestretch}%
{#1}\small\normalsize} \spacingset{1}

\title{Measurement error models: from nonparametric methods to deep neural networks}

\author{Zhirui Hu, Zheng Tracy Ke, and Jun S Liu\thanks{The authors gratefully acknowledge the support of the NSF grant DMS-1943902.}\\
{\it\small Department of Statistics, Harvard University}}
\maketitle

\begin{abstract}
The success of deep learning has inspired recent interests in applying neural networks in statistical inference. In this paper, we investigate the use of deep neural networks for nonparametric regression with measurement errors. 
We propose an efficient neural network design for estimating measurement error models, in which we use a fully connected feed-forward neural network (FNN) to approximate the regression function $f(x)$, a normalizing flow to approximate the prior distribution of $X$, and an inference network to approximate 
the posterior distribution of $X$.  
Our method utilizes recent advances in variational inference for deep neural networks, such as
the importance weight autoencoder, doubly reparametrized gradient estimator, 
and non-linear independent components estimation. 
We conduct an extensive numerical study to compare the neural network approach with classical nonparametric methods and observe  
that the neural network approach is more flexible in accommodating different classes of regression functions and performs superior or comparable to the best available method in nearly all settings.
\end{abstract}

\section{Introduction} \label{sec:intro}

The study of nonparametric regression with measurement error is a classical problem in statistics and has received a lot of attentions \citep{carroll2006measurement}. In a typical setting, the response $Y\in\mathbb{R}$ satisfies that $\mathbb{E}[Y|X]=f(X)$, where $X\in\mathbb{R}^d$ are the covariates and $f:\mathbb{R}^d\to \mathbb{R}$ is an unknown regression function. The covariates are observed with additive errors, i.e., we observe $W=X+U$ instead of $X$, where $U$ is a mean-zero random vector whose distribution is known. Given $(W, Y)$, the problem of interest is to estimate the nonparametric function $f$. 

Most existing methods for fitting measurement error models (MEMs) heavily use classical techniques in nonparametric statistics, such as kernel estimators, splines, and local polynomials. The deconvolution method \citep{Fan1993} combines the kernel estimator in nonparametric regression with the deconvolution technique in density estimation. Another class of methods approximate the regression function $f$ by splines and estimate the spline coefficients using estimating equations \citep{Jiang2018}. The simulation extrapolation method \citep{Carroll1999} starts from a standard nonparametric regression method (e.g., splines or local polynomials) and utilizes simulations to estimate and correct its bias when the covariates have errors. While these methods enjoy nice theoretical properties, they still face some major challenges in applications. The first issue is that each method is limited to a particular function class, such as nonparametric functions that are sufficiently smooth or functions that can be well approximated by splines. It is unclear how well they perform when the regression function $f$ is generated by a Gaussian process with a non-smooth kernel or when $f$ is a complicated function arising from scientific problems. Ideally, we hope to have a method that works for various function classes. Second, these methods require selecting critical tuning parameters, such as the bandwidth in a kernel estimator or the knots for splines. These tuning parameters can significantly affect the performance, but how to select them in a data-driven fashion is a hard problem. Especially, since the common bandwidth selection techniques were mainly designed for nonparametric regression without measurement errors, they may perform unsatisfactorily when the covariates have errors. The third challenge is generalization to multiple covariates (i.e., $d>1$). Most existing methods were only studied and evaluated for the case of $d=1$. Although some of them have extensions to $d>1$, the practical implementation can be inconvenient. For example, constructing a multivariate kernel estimator requires selecting the optimal bandwidth matrix, which is difficult in practice.

At the same time, the rapid growth of research on deep neural networks opens a new direction for addressing some difficult problems in classical statistics. Attempts have been made for density estimation \citep{liang2017well,singh2018nonparametric} and nonparametric regression \citep{schmidt2017nonparametric,bauer2019deep}. In this paper, we aim to integrate deep neural networks into the estimation of measurement error models. Some nice features of  the neural network make it promising for overcoming the challenges faced by classical nonparametric methods. First, it has been widely observed, empirically and theoretically, that deep neural networks have the ability of representing a variety of function classes \citep{barron1993universal,mhaskar1996neural,
maiorov2000near,lin2017does,rolnick2017power}, including many smooth function classes considered in classical nonparametric statistics. Hence, we can potentially use neural networks to develop a universal approach that works for all kinds of function classes. Second, deep neural networks tend to be resistant to overfitting and have great generalization power even when the parameter space has a very high complexity  \citep{golowich2018size,soudry2018implicit}. We note that, in classic nonparametric measurement error models, even a moderate $d$ significantly increases the complexity of parameter space, and thus may benefit significantly
from using neural networks. The tuning for neural networks is also less critical than the selection of smooth parameters (e.g., bandwidth) in nonparametric methods. For the latter, sub-optimal tuning parameters easily lead to overfitting or underfitting. For the former, the major tuning parameters are the architecture of the neural network. We shall use fully connected feed-forward neural networks (FNNs), which boils down to selecting the number of layers and the number of nodes for each layer. The resistance-to-overfitting by neural networks encourages us to set those numbers large without fine tuning.

In this article, we introduce our method of Neural Network for Measurement Error models (NNME), which uses a fully connected FNN to approximate the regression function $f(x)$, a normalizing flow \citep{tabak2013family} to approximate the prior distribution of $X$, and an inference network \citep{kingma2013auto,rezende2014stochastic} to approximate the posterior distribution of $X$. The training algorithm utilizes some recent advancements in variational inference methods for neural networks, particularly, the importance weighted autoencoder \citep{Burda2015} and the doubly reparametrized gradient estimator \citep{Tucker2018}. 
In fact, viewing the MEM as a latent variable model, our problem has a similar setting as the variational autoencoder (VAE). %\citep{kingma2013auto,rezende2014stochastic}. 
However, a direct application of VAE yields unsatisfactory performance. There is a nascent literature on improvements and alternatives of VAE \citep{Burda2015,roeder2017sticking,le2017auto,rainforth2018tighter,Tucker2018}, which is unfortunately not familiar to the statistics community. 
The description of our method also serves as introducing and elaborating these ideas to the statistics community.

We conducted an extensive numerical study to compare NNME with classical nonparametric methods for estimating MEMs. Although the neural network approach is promising, there is no guarantee that it will indeed outperform classical methods. A theoretical comparison is extremely difficult, as the theoretical understanding of deep learning is known to be challenging \citep{zhang2016understanding}. We thus focus on numerical comparison and hope to get a practical guideline of when the neural network method excels and to gain useful insight for future theoretical study. We investigate different function classes for $f$, including  smooth functions suitable for classical nonparametric methods, functions generated from Gaussian processes (such functions can be non-smooth), and  functions generated by some other unknown neural networks. We discover  that the neural network approach has a great flexibility in accommodating different function classes. It has reasonably good performance (sometimes, the best) in all the settings, while each competitor only works well for some specific function classes. In addition, the neural network approach is convenient to apply to the case of multiple covariates, but many classical nonparametric methods are difficult to implement for $d>1$.

The remaining part of this article is organized as follows: Section~\ref{sec:Method} introduces the NNME algorithm. Section~\ref{sec:Comparison} contains the comparison with other existing methods for MEMs. Section~\ref{sec:RealData} presents the application in two real datasets. Section~\ref{sec:Discuss} makes concluding remarks.

\nocite{fan2019selective}  

\section{The Neural Network for Nonparametric Regression with Measurement Errors} \label{sec:Method}

Let $Y\in\mathbb{R}$ be the response, and let $W\in\mathbb{R}^d$ be the vector of error-prone covariates. We assume
\begin{align}  \label{eqn:model}
& Y= f(X)+\epsilon, \qquad \epsilon \sim {\cal N}(0, \sigma^2)\;\; \mbox{and} \;\; {X\sim p_X(x),}\cr
& W = X + U,\qquad\;\;\; U\sim p_U(u), 
\end{align}
where $(X, U, \epsilon)$ are mutually independent, $p_X(x)$ is the (prior) distribution of error-free covariates, and $p_U(u)$ is the distribution of measurement errors satisfying that $\mathbb{E}U=0$.
Given $\{(w_i, y_i), i = 1,2, ..., n\}$ that are independent and identically distributed (IID) realizations of $(W, Y)$, the goal is to estimate the regression function $f$. 
We follow the convention to assume that the analytical form of the measurement error distribution $p_U(\cdot)$ is precisely known (in practice, this distribution is often estimated from other data source or determined by the prior knowledge). 
The distribution $p_X(\cdot)$ and the
variance of observation noise, $\sigma^2$, are unknown. 
Note that the assumption that $Y|X$ follows a normal distribution is only for convenience. It can be replaced by other parametric distributions, with minor modifications of our method.

\begin{figure}[!tb]
\centering
\includegraphics[width=0.9\textwidth]{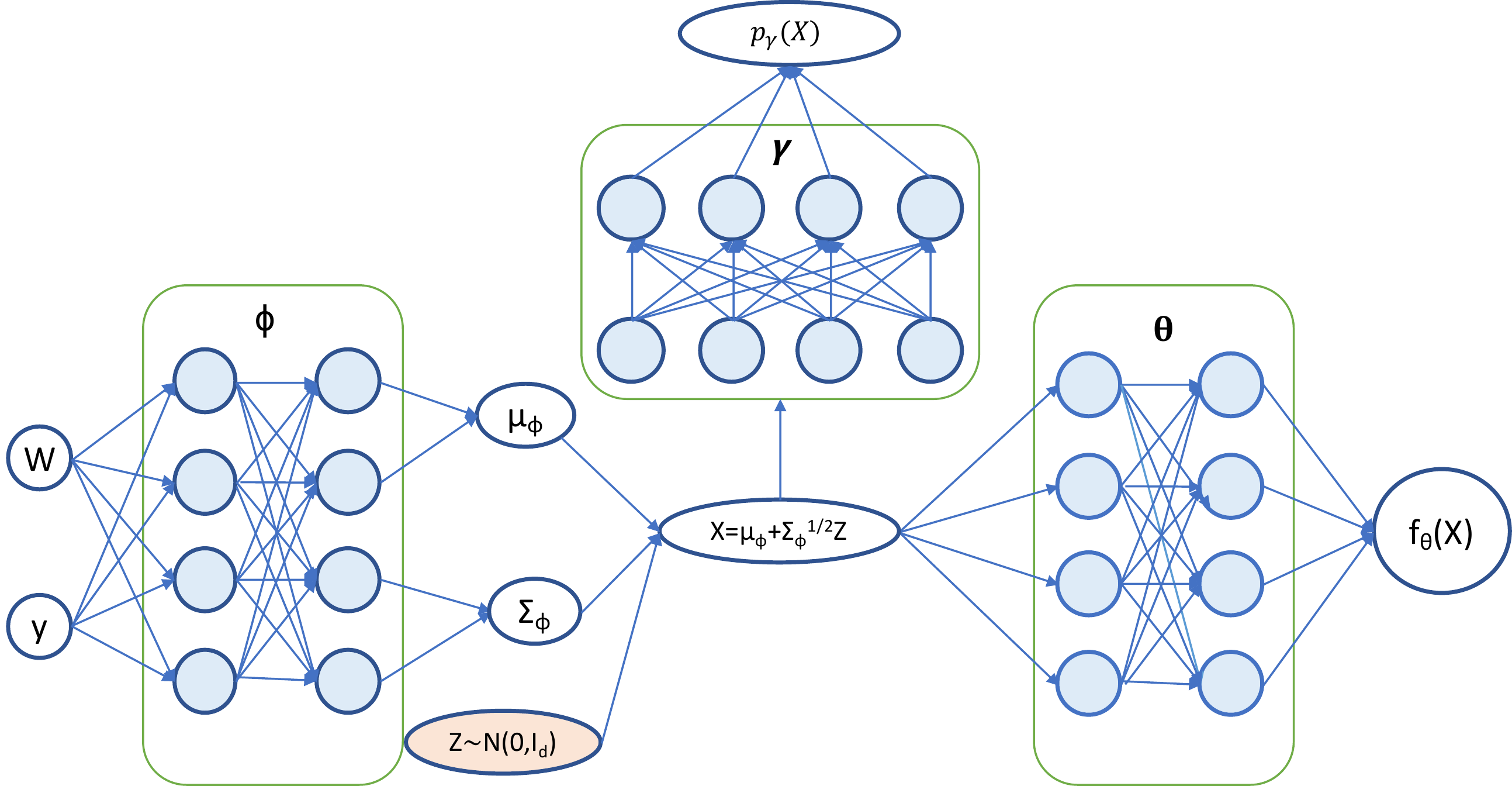}
\caption{A neural network structure for NNME. The input is $w$ and $y$, and the output is the estimated regression function $f_\theta(x)$. The {left} green block is an ``encoder,'' which consists of several fully connected layers with ReLU activation functions and the last layer with a linear function; the output of the encoder are parameters for the proposal distribution.  
The {right} green block is a ``decoder,'' which has the same network structure as the encoder; the input are random samples of $x$, and the output are estimated values of $f_\theta(x)$. {The top green block is another ``decoder,'' which consists of a few coupling layers of a normalizing flow; the input are random samples of $x$, and the output is the estimated marginal density of $X$. }}\label{fig:nn2} 
\end{figure}

\subsection{The neural network structure} \label{subsec:NNstructure}
We use a fully connected feed-forward neural network (FNN) to model the  regression function $f$. With $\theta$ denoting parameters of this FNN, we can write $f=f_\theta$. Next, we use a normalizing flow \citep{tabak2013family} to represent the prior distribution $p_X$. A normalizing flow is a sequence of transformations $g_1,g_2,\ldots,g_m$, where each $g_j$ is an invertible mapping from $\mathbb{R}^d$ to $\mathbb{R}^d$. Let $V\in\mathbb{R}^d$ be a random vector whose density has a simple analytic form (e.g., $V\sim {\cal N}(0, I_d)$). We model the density of $X$ by assuming $X=g^{-1}(V)$, where $g=g_m\circ g_{m-1}\circ\ldots \circ g_1$. Let $J(v)$ denote the Jacobian of the mapping $g$. The change of variables formula implies that
\[
p_X(x) = p_V(v) \cdot | \det J(v)|\Big|_{v=g(x)}. 
\]
By parametrizing each $g_j$ via neural network layers, this idea has been applied to density estimation \citep{papamakarios2017masked} and variational inference \citep{rezende2015variational}. We adopt the specific flow proposed by \cite{dinh2014nice}, called non-linear independent components estimation (NICE). NICE parametrizes each invertible transformation $x = g_j(v)$ as follows: $x_{I_1}=v_{I_1}$ and $x_{I_2}=v_{I_2}+h(v_{I_1})$, where $I_1\cup I_2$ is a partition of $\{1,2,\ldots,d\}$ and $h$ is a neural network with $|I_1|$ input units and $|I_2|$ output units. Such a mapping has a trivial inverse: $v_{I_1}=x_{I_1}$ and $v_{I_2}=x_{I_2}-h(x_{I_1})$. Furthermore, its Jacobian always has a unit determinant, regardless of the choice of $h$. It thus follows that  
\begin{equation} \label{eqn:densityX}
p_X(x) = p_{V}(g_{\gamma}(x)),  \qquad\mbox{where $\gamma$ denotes parameters of NICE}. 
\end{equation}
Combining \eqref{eqn:densityX} with the MEM in \eqref{eqn:model}, we can write down the joint density of $(W, Y, X)$:
\begin{equation} \label{eqn:joint-density}
p(w, y, x; \theta, \gamma,\sigma^2) = p_{V}(g_{\gamma}(x))\cdot p_{U}(w-x)\cdot p_{\epsilon}(y-f_\theta(x); \sigma^2). 
\end{equation}
Here, $p_U(\cdot)$ is the density of measurement errors, which is known; $p_{V}(\cdot)$ is a simple analytical density, which is typically chosen as  the $d$-dimensional standard Gaussian; 
and $p_{\epsilon}(\cdot)$ is the normal density of noise in $y$. For a moment, we assume $\sigma^2$ is known and write this joint density as $p_{\theta,\gamma}(w, y, x)$. Later, in our main algorithm, $\sigma^2$ will be estimated jointly with other parameters.

Now that the MEM is  parametrized by $(\theta,\gamma)$, where $\theta$ contains parameters of interest and $\gamma$ contains nuisance parameters, we estimate $(\theta,\gamma)$ by maximizing the marginal log-likelihood of the observed variables:
\begin{equation} \label{eqn:marLR}
L^*(\theta, \gamma |w,y)\equiv \log \int  p_{\theta,\gamma}(w,y,x)dx. 
\end{equation}
This objective function involves an integral over $x$ and cannot be evaluated analytically.

The variational inference is a common tool for parameter estimation and inference in such latent variable models \citep{jordan1999introduction}.
The variational auto-encoder (VAE) \citep{kingma2013auto,rezende2014stochastic} is a variational inference approach via neural networks. Standard variational inference approaches approximate the intractable posterior distribution by a parametric family whose densities have simple analytical forms, whereas VAE achieves the same goal %approximates the intractable posterior 
by an inference network (also called a recognition model), where the posterior distribution is approximated by a neural network transformation of multivariate normal densities. It has been widely recognized that VAE is capable of approximating complex posterior distributions. 
\cite{Burda2015} proposed an improvement of VAE by incorporating the idea of importance sampling, namely the importance weighted autoencoder (IWAE). We adapt IWAE to maximize the marginal log-likelihood in \eqref{eqn:marLR}.

Let $x_1,x_2,\ldots,x_K$ be IID samples drawn from a proposal distribution $q(x|w,y)$. According to Jensen's inequality, the marginal log-likelihood has a lower bound: 
\begin{align*}
L^*(\theta, \gamma |w,y) &= \log\left(  \mathbb{E}_{x_{1:K}\sim q(\cdot |w,y)}\left[ \frac{1}{K}\sum_{k=1}^K\frac{p_{\theta,\gamma}(w,y,x)}{q(x_k|w,y)}\right] \right) \cr
&\geq \mathbb{E}_{x_{1:K}\sim q(\cdot |w,y)}\left[ \log\left( \frac{1}{K} \sum_{k=1}^K \frac{p_\theta(w,y,x_k)}{q(x_k|w,y)}\right) \right] \; \equiv L(\theta,\gamma|w,y). 
\end{align*}
This lower bound {$L(\theta,\gamma|w,y)$ is called an evidence lower bound (ELBO).} It is tighter if $q(x|w,y)$ is closer to the true posterior distribution of $X$ and/or $K$ is larger. 

The idea of VAE is to use a neural network to model the proposal distribution and optimize it jointly with other parameters. However, VAE only corresponds to $K=1$. One can use the idea of importance sampling (e.g., see \cite{liu2008monte}) with an appropriate sample size $K$ to tighten the lower bound. IWAE of \cite{Burda2015} combines these two ideas.
We adopt IWAE and model $q(x|w,y)$ as 
a multivariate normal distribution ${\cal N}\bigl( \mu_\phi(w,y), \Sigma_\phi(w, y) \bigr)$, where the vector $\mu_\phi(w,y)\in\mathbb{R}^d$ and the diagonal matrix $\Sigma_\phi(w,y)\in \mathbb{R}^{d\times d}$  are both determined by another FNN {whose parameters are denoted by $\phi$}. This is equivalent to drawing samples $x_k$ from
\begin{equation} \label{eqn:proposal}
x(\phi, z_k) = \mu_{\phi}(w,y) +[\Sigma_{\phi}(w,y)]^{1/2}z_k, \qquad \mbox{where}\quad z_1,z_2,\ldots,z_K\overset{iid}{\sim}{\cal N}(0, I_d).  
\end{equation}
{Let $q_{\phi}(x|w,y)$ denote the density of ${\cal N}\bigl( \mu_\phi(w,y), \Sigma_\phi(w, y) \bigr)$ and let $x(\phi, z_k)$ be the mapping in \eqref{eqn:proposal}.}
We consider the following objective:
\begin{equation}     \label{eqn:obj}
Q(\theta,\gamma, \phi|w,y)\equiv   \mathbb{E}_{z_{1:K}\sim {\cal N}(0, I_d)}\left[ \log\left(\frac{1}{K}\sum_{k=1}^K\frac{p_{\theta,\gamma}\big(w, y, x(\phi, z_k)\big)}{q_\phi\big(x(\phi, z_k) |w, y\big)}\right)\right]. 
\end{equation}
The estimator is 
\begin{equation} \label{eqn:estimator}
\bigl(\hat{\theta}(w,y),\; {\hat{\gamma}(w,y)}, \; \hat{\phi}(w,y)\bigr) = \mathrm{argmax}_{\theta,\gamma, \phi} Q(\theta,\gamma, \phi|w,y). 
\end{equation}

The neural network structure that implements this estimator is shown in Figure~\ref{fig:nn2}. The FNN parametrized by $\phi$ that generates samples from the proposal distribution is called ``encoder.'' The FNN parametrized by $\theta$ that approximates the regression function $f(x)$ { and the normalizing flow parametrized by $\gamma$ that approximates the prior distribution of $X$} are both called ``decoders.'' The ``encoder'' takes data $(W,Y)$ as input, and outputs parameters $(\mu_\phi, \Sigma_\phi)$ for the proposal distribution. The ``decoders'' take as input the random samples generated from the proposal distribution, and outputs the estimated regression function $f_\theta(x)$ and density $p_{\gamma}(x)$ at those random samples of $X$.

\subsection{The gradient ascent algorithm} \label{subsec:GD}
{We  use a gradient ascent algorithm to solve \eqref{eqn:estimator}. Given initial values $(\theta^{(0)}, \gamma^{(0)}, \phi^{(0)})$, the update at iteration $t$ is}
\begin{equation}
(\theta^{(t+1)},\gamma^{(t+1)}, \phi^{(t+1)})=(\theta^{(t)},\gamma^{(t)}, \phi^{(t)}) -\alpha_t\cdot   \nabla Q(\theta^{(t)},\gamma^{(t)}, \phi^{(t)}|w,y), \label{gradient_ascent}
\end{equation}
where $\alpha_t>0$ is {the step size at iteration $t$,} often called the learning rate. {We use the adaptive learning rate suggested by \cite{kingma2014adam}, which is  $\alpha_t=\alpha_0\sqrt{1-\alpha_2^t}/(1-\alpha_1^t)$, for some default parameters $(\alpha_0, \alpha_1, \alpha_2)$.}

{It remains to compute the gradient. Write for short $\beta_k =\frac{p_{\theta,\gamma}(w, y, x(\phi, z_k))}{q_\phi(x(\phi, z_k) |w, y)}$, $1\leq k\leq K$. By direct calculations, we have
\begin{equation} \label{eqn:gradient}
\nabla_{\theta,\gamma, \phi}Q(\theta,\gamma, \phi|w,y)=\mathbb{E}_{z_{1:K}\sim {\cal N}(0, I_d)}\left[\sum_{k=1}^K \frac{\beta_k}{\sum_{\ell=1}^K \beta_\ell} \nabla_{\theta, \gamma, \phi} \log \left(\frac{p_{\theta,\gamma}(w, y, x(\phi, z_k))}{q_\phi(x(\phi, z_k) |w, y)}\right)\right].  
\end{equation}
The Monte Carlo method approximates the gradients by replacing the expectation in \eqref{eqn:gradient} by realized values at a random sample of $z_{1:K}$.}
At each iteration, we draw $z_1,z_2,\ldots,z_K$ IID from ${\cal N}(0, I_d)$ and estimate the gradient \eqref{eqn:gradient} by
\begin{equation} \label{eqn:plainGD}
\widehat{\nabla_{\theta,\gamma,\phi}Q}(\theta,\gamma, \phi|w,y) = \sum_{k=1}^K \frac{\beta_k}{\sum_{\ell=1}^K \beta_\ell} \nabla_{\theta,\gamma, \phi} \log \left(\frac{p_{\theta,\gamma}(w, y, x(\phi, z_k))}{q_\phi(x(\phi, z_k) |w, y)}\right). 
\end{equation}

This plain gradient estimator works for approximating $\nabla_\theta Q$ and $\nabla_{\gamma}Q$, but not $\nabla_\phi Q$, where we recall that $\phi$ contains parameters in the encoder. The issue of $\widehat{\nabla_{\phi}Q}$ happens when $K\to\infty$. By theory of IWAE, the objective in \eqref{eqn:obj}, as a lower bound of the true marginal log-likelihood, converges to the true marginal log-likelihood as $K\to\infty$ \citep{Burda2015}. Therefore, a larger $K$ (i.e., more ``importance samples'') should be favored.
Unfortunately, if \eqref{eqn:obj} is solved by a plain gradient ascent algorithm, the performance is often worse with an increased $K$. The reason is that $\widehat{\nabla_\phi  Q}$ is too ``noisy.'' The true gradient $\nabla_\phi Q$ vanishes as $K\to\infty$; therefore, it is the bias $(\mathbb{E}_{z_{1:K}}[\widehat{\nabla_{\phi}Q}]-\nabla_\phi Q)$, not the true gradient $\nabla_\phi Q$, that contains the signal. Since the bias diminishes faster than the standard deviation, as $K$ increases, the signal-to-noise ratio in the gradient actually decreases. This is, however, not a problem for the gradient with respect to $(\theta,\gamma)$, as the true gradient $\nabla_{\theta,\gamma} Q$ does not vanish as $K\to\infty$. See \cite{rainforth2018tighter} for a rigorous theoretical justification.

{\cite{Tucker2018} proposed a method to resolve the above issue. They derived an alternative expression of the gradient with respect to $\phi$: 
\begin{align} \label{eqn:gradient2}
 & \nabla_{\phi}Q(\theta,\gamma, \phi|w,y)\cr
=\; & \mathbb{E}_{z_{1:K}\sim {\cal N}(0, I_d)}\left[\sum_{k=1}^K \biggl(\frac{\beta_k}{\sum_{\ell=1}^K \beta_\ell}\biggr)^2\cdot \frac{\partial}{\partial x_k} \log \left(\frac{p_{\theta,\gamma}(w, y, x_k)}{q_\phi(x_k |w, y)}\right)\bigg|_{x_k=x(\phi, z_k)}\cdot \frac{\partial}{\partial \phi} x(\phi, z_k)\right], 
\end{align} 
where $\beta_k$ is the same as in \eqref{eqn:gradient}. {We note that \eqref{eqn:gradient2} cannot be derived from \eqref{eqn:gradient} via a simple chain rule. In fact, a direct use of the chain rule should yield a term related to $\frac{\partial}{\partial \phi} q_{\phi}(x_k|w,y)$, but there is no such term in \eqref{eqn:gradient2}. This alternative gradient formula motivates another Monte Carlo estimator for $\nabla_\phi Q$:}
\begin{equation} \label{eqn:DReG}
\widetilde{\nabla_{\phi}Q}(\theta,\phi|w,y)=\sum_{k=1}^K \biggl(\frac{\beta_k}{\sum_{\ell=1}^K \beta_\ell}\biggr)^2\cdot \frac{\partial}{\partial x_k} \log \left(\frac{p_{\theta,\gamma}(w, y, x_k)}{q_\phi(x_k |w, y)}\right)\bigg|_{x_k=x(\phi, z_k)}\cdot \frac{\partial}{\partial \phi} x(\phi, z_k), 
\end{equation}
which is called the doubly reparametrized gradient estimator. Compared with the plain gradient estimator \eqref{eqn:plainGD}, the variance of this estimator converges to zero at a faster rate, so that the signal-to-noise ratio still increases with $K$ (see \cite{Tucker2018} for a detailed explanation). The difference between \eqref{eqn:plainGD} and \eqref{eqn:DReG} is analogous to the difference between the score function gradient estimator (or reinforce gradient estimator) and the reparametrization gradient estimator in Monte Carlo variational inference \citep{kingma2013auto,titsias2014doubly,rezende2014stochastic}. It is well-known that the reparametrization gradient estimator has a smaller variance.

\subsection{Our method: NNME} \label{subsec:NNME}
Our main algorithm, called \textbf{N}eural \textbf{N}etwork for \textbf{M}easurement \textbf{E}rror (NNME), is summarized in Algorithm 1.

\begin{algorithm}[!htb]
\caption{NNME} \label{algo:overall}
\textbf{Data}: Training data $\{w_i, y_i\}_{i=1}^n$ and prespecified values $\{x_i\}_{i=1}^m$ for evaluating $f_{\theta}(x)$.\\
\textbf{Input}: The density $p_U(\cdot)$ of measurement error, 
the neural network structure, parameters $(\lambda_0,\lambda_1,\lambda_2)$, maximum number of epochs $\mathrm{Max}\_\mathrm{Epoch}$, and number of importance samples $K$.
\begin{enumerate}
\item Pre-processing: Standardize $y$ and each covariate of $w$.
\item Initialization:
            For each of the two networks representing $f_{\theta}$ and $g_{\gamma}$, pre-train it using $\{w_i, y_i\}_{i=1}^n$ as if there is no measurement error. In pre-training,
            %The network structure is specified by $T$, and 
            the loss function is the mean squared error (MSE) plus an L2 penalization term, $\lambda_0\|\theta\|^2$ or $\lambda_2\|\gamma\|^2$. Initialize $(\theta,\gamma)$ from pre-training; initialize $\phi$ randomly; and initialize $\sigma^2$ by the MSE from pre-training.
            %\item  \textbf{Training of the neural network:} 
            \item For epoch = 1 to Max\_Epoch, run the following steps to obtain $(\hat{\theta},\hat{\gamma},\hat{\phi})$:
            \begin{itemize}
                \item Draw Monte Carlo samples $\{z_{ik}\}_{1\leq i\leq n, 1\leq k\leq K}$ IID from ${\cal N}(0,I_d)$. 
                \item Compute the gradients $\widehat{\nabla_{\theta}Q}$, $\widehat{\nabla_{\gamma}Q}$, and $\widetilde{\nabla_{\phi}Q}$ as in \eqref{eqn:NNMEgradients}. Let $\widehat{\nabla_{\theta}Q}_*=\widehat{\nabla_{\theta}Q}+2\lambda_0\theta$, $\widehat{\nabla_{\gamma}Q}_*=\widehat{\nabla_{\gamma}Q}+2\lambda_2\gamma$ and $\widetilde{\nabla_{\phi}Q}_*=\widetilde{\nabla_{\phi}Q}+2\lambda_1 \phi$ (to account for the L2 penalty). Update $(\theta,\gamma, \phi)$ via gradient ascent as in \eqref{gradient_ascent}.   
                \item Update $\sigma^2$ by (\ref{eqn:sy}) using the training samples.
                \end{itemize}
\end{enumerate}
\textbf{Output}: The estimated values $\{f_{\hat{\theta}}(x_i)\}_{i=1}^m$. 
\end{algorithm}

We now provide a more detailed explanation of NNME.
Note that the calculations in Sections~\ref{subsec:NNstructure}-\ref{subsec:GD} are for $n=1$. For a general $n$, write $w^{(n)}=\{w_i\}_{i=1}^n$ and $y^{(n)}=\{y_i\}_{i=1}^n$. At each epoch, we draw samples $\{z_{ik}\}_{1\leq i\leq n,1\leq k\leq K}$ IID from ${\cal N}(0, I_d)$ and obtain $x_{ik}=x(\phi,z_{ik})$. Similar to \eqref{eqn:plainGD} and \eqref{eqn:DReG}, we have
\begin{align} \label{eqn:NNMEgradients}
\widehat{\nabla_{\theta,\gamma}Q}(\theta,\gamma,\phi|w^{(n)},y^{(n)}) & = \sum_{i=1}^n \sum_{k=1}^K \frac{\beta_{ik}}{\sum_{\ell=1}^K \beta_{i\ell}} \nabla_{\theta,\gamma} \log \left(\frac{p_{\theta,\gamma}(w_i, y_i, x(\phi, z_{ik}))}{q_\phi(x(\phi, z_{ik}) |w_i, y_i)}\right),\\
\widetilde{\nabla_{\phi}Q}(\theta,\gamma,\phi|w^{(n)},y^{(n)})&=\sum_{i=1}^n \sum_{k=1}^K \biggl(\frac{\beta_{ik}}{\sum_{\ell=1}^K \beta_{i\ell}}\biggr)^2 \left[\frac{\partial}{\partial x_{ik}} \log \left(\frac{p_{\theta,\gamma}(w_i, y_i, x_{ik})}{q_\phi(x_{ik} |w_i, y_i)}\right)\right] \frac{\partial}{\partial \phi} x(\phi, z_{ik}), \nonumber
\end{align}
where $\beta_{ik}=p_{\theta,\gamma}(w_i, y_i, x_{ik})/q_\phi(x_{ik}|w_i, y_i)$ and the expressions of 
$p_{\theta,\gamma}(w,y,x)$ and $q_\phi(x|w,y)$ are given by \eqref{eqn:joint-density} and \eqref{eqn:proposal}, respectively. The gradient computation involves calculation of $\nabla_\theta f_\theta$, $\nabla_{\gamma}g_{\gamma}$, and $(\nabla_\phi\mu_\phi, \nabla_\phi\Sigma_\phi)$, which are computed via the back propagation algorithm \citep{mcclelland1986parallel,hecht1992theory}. We also use the $L_2$-regularization trick to reduce numerical instability. We add a penalty $\lambda_0\|\theta\|^2+\lambda_1\|\phi\|^2 +\lambda_2\|\gamma\|^2$ to the objective function. It is similar in spirit to the ridge regression and helps stabilize the numerical performance. This $L_2$ penalty changes nothing of the gradient ascent algorithm except for an additional term to each of the gradients $(\nabla_\theta, \nabla_\phi, \nabla_{\gamma})$; 
see Step 3 of Algorithm 1.
The noise variance $\sigma^2$, which is assumed known in derivations of Sections~\ref{subsec:NNstructure}-\ref{subsec:GD}, can be estimated along with training the neural network. 
At each epoch, we estimate $\sigma^2$ by a weighed sum of residuals:
\begin{equation} \label{eqn:sy}
    \hat\sigma^2 = \frac1{nK}\sum_{i=1}^n\sum_{k=1}^K (y_i - f_\theta(x_{ik}))^2 \frac{\beta_{ik}}{\sum_{\ell=1}^K \beta_{i\ell }}. 
\end{equation} 
No additional computing effort is needed, since $\beta_{ik}$ are already obtained in the computation of gradients.

Regarding the input of Algorithm 1, the measurement error density $p_U(\cdot)$ is supplied by the user, as in other measurement error methods. In the literature, it is common to assume that the measurement errors follow ${\cal N}(0, \sigma_0^2I_d)$, where $\sigma_0^2$ is estimated from other data source or determined by prior knowledge. 
%The prior distribution $p_1(x)$ for $X$ is often set as a uniform distribution or a $t$-distribution. 
By default, we set the number of Monte Carlo samples as $K=50$ and the maximum number of epochs as Max\_Epoch = 500. The neural network structure $T$ and the L2-penalty coefficients $(\lambda_0,\lambda_1,\lambda_2)$ are chosen by a 5-fold cross validation. The validation loss is calculated as follows: For each test sample $(w_i, y_i)$, we draw $\{z_{ik}\}_{k=1}^K$ IID from ${\cal N}(0, I_d)$ and compute $x_{ik}$ and $\beta_{ik}$ by plugging in the estimated parameters $(\hat{\theta},\hat{\gamma}, \hat{\phi})$ and the testing $(w_i, y_i)$. These numbers are then plugged into \eqref{eqn:sy} to give the validation loss.  
%$\sum_i (y_i-f_\theta(x_i))^2$. 
The implementation of gradient ascent is via Adam \citep{kingma2014adam}, with batch size equal to $\min\{512, n\}$. Adam requires three parameters $(\alpha_0,\alpha_1,\alpha_2)$ to determine the learning rate (see the paragraph below \eqref{gradient_ascent}), where we set as the default values as $(0.001, 0.9, 0.999)$.

The method we propose here uses three neural networks, one for representing the regression function $f$, one for representing the (prior) distribution of $X$, and one for approximating the posterior distribution of $X$. If the prior distribution of $X$ is less complicated, we can model it with a parametric density $p_{\gamma}(x)$, where $\gamma$ now has a different meaning and represents parameters of the parametric density. Despite that the expression of $p_{\theta,\gamma}(w,y,x)$ changes, the objective \eqref{eqn:obj} and the gradient ascent algorithm remain the same. It gives rise to a simplified version of NNME. This version works well if either the dimension is $1$ and/or the sample size is not very large. In such cases, using a parametric model for the distribution of $X$ leads to more stable numerical performance, even if the parametric model is misspecified (see Section~\ref{subsec:otherversion}). Throughout this paper, we still call this simplified version NNME but will clarify whether a neural network or parametric model is used for the distribution of $X$.  
}

%%%%%%%%%%%%%%%%%%%%%%%%%%
\subsection{Comparison with alternative neural network algorithms} \label{subsec:otherversion}
We compare NNME with alternative options of neural network algorithms for  MEMs. 
In the first method (``NN''), we regress $y$ directly on $w$ using FNN, pretending that there is no measurement error. The second method is called the ``Maximizing joint likelihood'' method (``MJL''), which maximizes the joint log-likelihood of $(w, x, y)$. We use an FNN to approximate the MLE of $x$ given $(w, y)$ and another FNN to approximate the function $f(x)$. In comparison with NNME, the MJL approach is analogous to ``optimal imputation,'' which does not  account for uncertainties in the imputation of $x_i$; see Appendix~\ref{app:other-methods} for more details. In the third method (``VAE''), 
%we use a variational distribution to approximate the posterior distribution of $x$ 
we apply the standard variational autoencoder, which corresponds to $K=1$ in the objective \eqref{eqn:obj}. To approximate the objective function at each iteration, the VAE requires multiple Monte Carlo samples, which are denoted as  $z_1,\ldots,z_L\sim {\cal N}(0, I_d)$. Then, the objective in VAE is approximated by 
\begin{equation}  \label{ISE}
\frac{1}{L}\sum_{\ell=1}^L \log\left(\frac{p_{\theta,\gamma}(w, y, x(\phi, z_\ell))}{q_\phi(x(\phi, z_\ell) |w, y)}\right). 
\end{equation}
The implementation details are in Appendix~\ref{app:other-methods}. The fourth method (``GA'', abbreviation for gradient ascent) maximizes \eqref{eqn:obj} using the plain gradient estimator in \eqref{eqn:plainGD}. The last is our algorithm (``NNME''), where we maximize \eqref{eqn:obj} with the doubly reparametrized gradient estimator. For all methods, we add L2 regularization on the weights in neural networks.

\begin{table}[!hbt]
\centering
\caption{Alternative neural network algorithms.} \label{tb:NN-algs}
\scalebox{.95}{\begin{tabular}{c | l} %|c|c|c|c|c|c|c|
 \toprule
 Abbreviation & Description of algorithms\\
 \hline
NN & using FNN to fit a nonparametric regression directly on $(w, y)$\\
MJL & maximizing the joint log likelihood of $(w,x,y)$\\
VAE & approximating the marginal log likelihood by variational autoencoder \\
GA & solving \eqref{eqn:obj} by the plain gradient ascent\\
\hline
NNME & solving \eqref{eqn:obj} with the doubly reparameterized gradient estimator \\
\hline
\end{tabular}}
\end{table}

Below, we use two numerical examples to compare different neural network algorithms. We measure the performance by integrated squared error (ISE):
\[
\int [f_{\hat{\theta}}(x) - f_\theta(x)]^2 dx. 
\]
The three methods, VAE, GA and NNME, all require modeling of the prior distribtion on $X$. For a fair comparison, we use the same prior for all three methods, which is $2\cdot t_3$, where $t_3$ denotes a $d$-variate distribution whose each coordinate is an independent $t$-distribution with 3 degrees of freedom. We also use the same cross-validation procedure for selecting tuning parameters (including the neural network structure) in all the five methods.

\medskip

{\bf Experiment 1: The univariate case}. In this experiment, the true regression function is $f(x) = sin(\pi x)$. The data are generated from \eqref{eqn:model}, where $x_i$'s are drawn from $unif(-2, 2)$, the measurement error distribution is $N(0, \sigma_0^2)$, and the noise standard deviations are $\sigma=\sigma_0=0.1$. The parameter $\sigma_0$ is supplied to all algorithms. 

\begin{figure}[!tb]
\centering
\begin{subfigure}[b]{1\textwidth}
\includegraphics[width=1.05\textwidth]{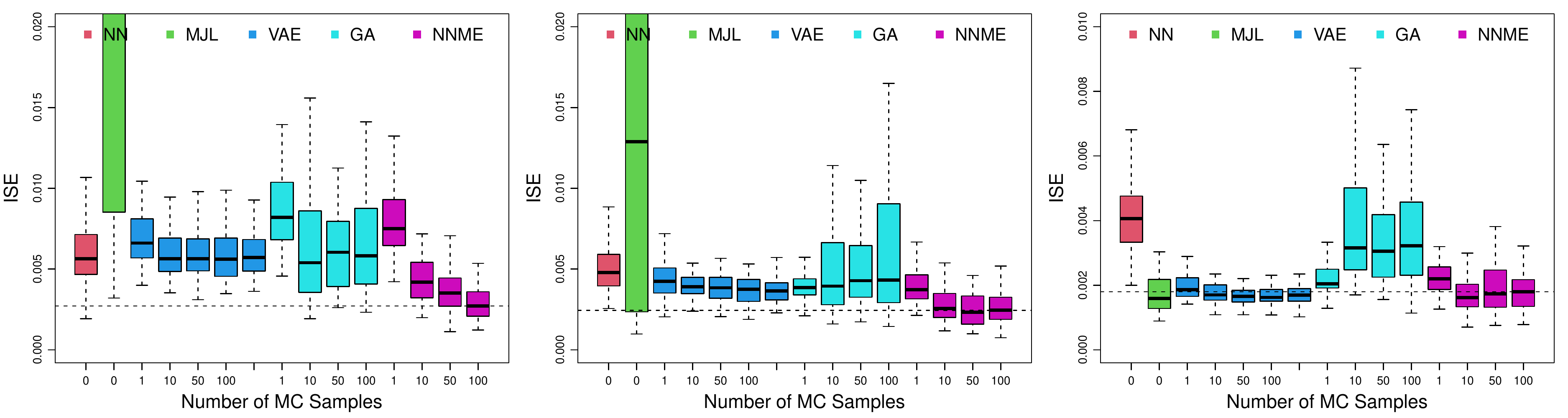}
\caption{The univariate example, where $f(x) = \sin(\pi x)$. From left to right, the sample size is $n=1000, 2000, 5000$. }
\end{subfigure}
\begin{subfigure}[b]{1\textwidth}
\includegraphics[width=1.05\textwidth]{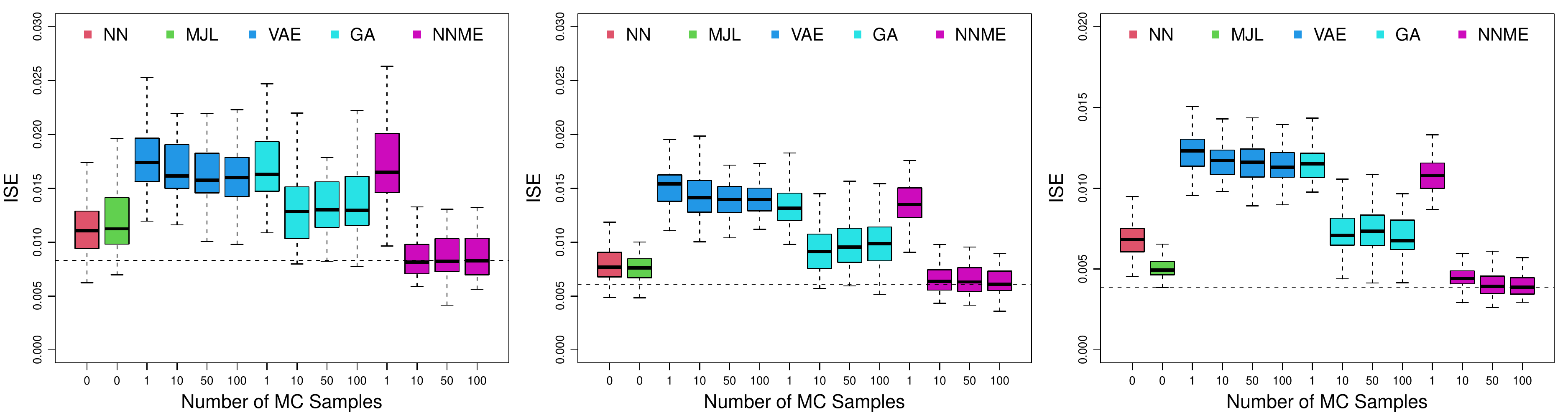}
\caption{The trivariate example, where $f(x)=(x_1x_2 + x_3)^2$. From left to right, the sample size is $n=1000, 2000, 5000$. }
\end{subfigure}
\caption{Comparison of different neural network algorithms. Y-axis: box plots of ISE based on 50 repetitions, X-axis: number of Monte Carlo samples at each iteration (this number is shown as 0 if the method does not use MC samples). 
For illustration, each figure contains a dash line, which is the median ISE for NNME with 100 Monte Carlo samples. }\label{fig:simu1_mise} 
\end{figure}

The results are shown in Figure~\ref{fig:simu1_mise} (a). From left to right, the sample size is $1000$, $2000$, and $5000$, respectively. Each of the three methods, VAE, GA and NNME, requires multiple Monte Carlo samples at each iteration. Note that 
\begin{align*}
& \mbox{number of MC samples}\cr
&=\begin{cases}
K \mbox{ (number of importance samples)}, & \mbox{for GA and NNME},\\
L \mbox{ (number of repeated draws at each iteration)}, & \mbox{for VAE},
\end{cases} 
\end{align*}
We have tried $K\in\{1, 10,50,100\}$ and $L\in \{1,10,50,100,200\}$.

There are a few noteworthy observations. First, NN always has a higher error than the best of other methods. It suggests that ignoring measurement errors yields unsatisfactory performance. Additionally, MJL incurs a huge error when the sample size is $1000$ or $2000$, suggesting that the maximum joint likelihood approach only works for sufficiently large sample sizes (e.g., $n\geq 5000$). Second, we compare the three algorithms: VAE, GA and NNME, all aiming to maximize an ELBO of the marginal log-likelihood. The VAE objective is based on only one importance sample (i.e., $K=1$). It has a similar performance with NNME for $n=5000$, but its performance is worse than that of NNME for $n\in \{1000,2000\}$. Additionally, increasing $L$, the number of repeated draws at each iteration, yields no significant improvement. This suggests that IWAE (i.e., $K>1$) is indeed a better option. GA and NNME use different gradient ascent algorithms to solve the same IWAE objective. For $K=1$, the two have similar performances, but for larger $K$, GA is much worse than NNME. This shows that, without the doubly reparametrized technique in gradient estimation, the advantage gained by increasing the number of Monte Carlo samples is counterbalanced by the large variance in gradients. In contrast, increasing $K$ in NNME significantly improves the performance. For $n\in\{2000,5000\}$, NNME with $100$ importance samples achieves the best performance and is much better than the  alternatives. For $n=5000$, MJL and VAE also perform well, and NNME performs similarly to them.  

In Appendix~\ref{app:other-methods}, we also investigate the estimated curve $f_{\hat{\theta}}(x)$ and the fitted values $\hat{x}$ for all methods, which help explain the reported ISEs. We relegate details to the appendix.

\medskip

{\bf Experiment 2: The multivariate case}. In this experiment, the true regression function is $f(x) = (x_1x_2 + x_3)^2$. We generate $\{x_{ij}\}_{1\leq i\leq N, 1\leq j\leq 3}$ IID from $unif(-1, 1)$, and let the measurement error distribution be ${\cal N}(0, \sigma_0^2 I_3)$, with $\sigma = \sigma_0 = 0.1$. Other settings are similar to those in Experiment 1.

The results are shown in Figure~\ref{fig:simu1_mise} (b). Compared with the univariate case, the performance of VAE is much worse because the ELBO of the marginal log-likelihood used by VAE becomes less tight for dimension higher than $1$. In \eqref{eqn:proposal}, $\Sigma_\phi$ is set as a diagonal matrix, implying that  a product measure is used to approximate the posterior distribution of $X$, which is restrictive for $d>1$. The  two IWAE based methods, GA and NNME, help tighten the lower bound by increasing the number of importance samples, and perform better than VAE. In comparison, NNME is significantly better than GA due to the use of doubly reparametrized gradient estimator; this is the same as what we have observed in Experiment 1. 

\medskip

\begin{figure}[!htb]
\centering
\includegraphics[width=0.99\textwidth, height=.51\textwidth]{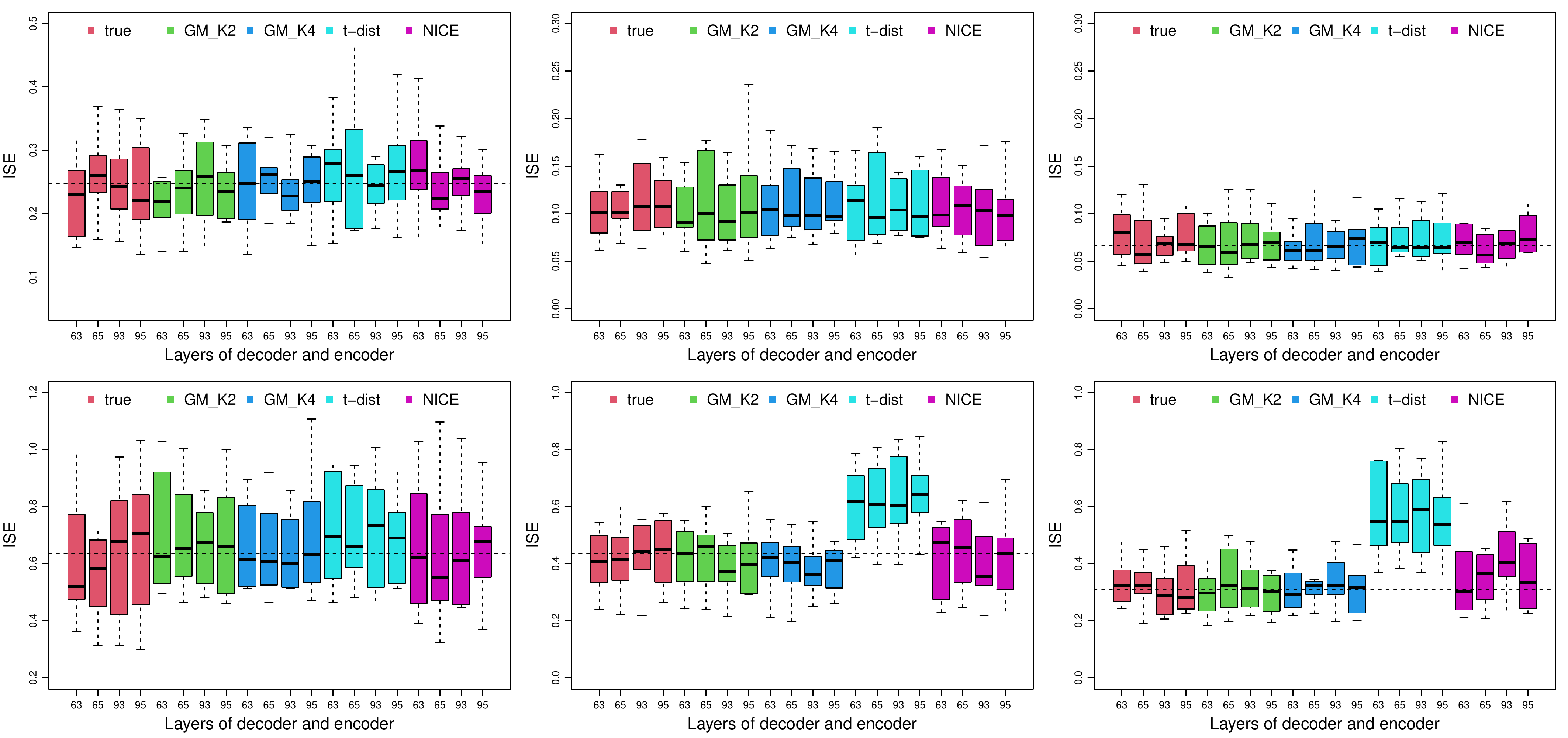}
\caption{Comparison of different versions of NNME where the assumed model for $X$ varies. Y-axis: box plots of ISE based on 10 repetitions; X-axis: the neural network structure (see the main text). In the top and bottom panels, the standard deviation of measurement errors is 0.05 and 0.2, respectively; from left to right, the sample size is 1000, 4000, 8000, respectively. 
For illustration, each plot has a dash line, which is the median ISE when the true distribution of $X$ is used (median is calculated among all repetitions in four network structures).}\label{fig:learnX} 
\end{figure}

{{\bf Experiment 3: Variants of NNME}. In this experiment, we investigate a few variants of NNME that use different models for the prior (marginal) distribution of $X$. The true regression function is generated from a Gaussian process on $\mathbb{R}^2$ (see Appendix~\ref{subsec:LearnXdist} for details). We simulate $\{x_i\}_{i=1}^n$ IID from a 2-component Gaussian mixture distribution:
\begin{equation}\label{simu-X-distribution}
x_i \overset{iid}{\sim}0.7 \cdot {\cal N}\left(\begin{bmatrix}-0.4 \\ 0.2\end{bmatrix}, \; \begin{bmatrix}0.2^2 &0 \\
0 & 0.3^2
\end{bmatrix}\right) + 0.3 \cdot {\cal N}\left(\begin{bmatrix}0.2 \\ 0.4\end{bmatrix},\;  \begin{bmatrix}0.3^2 &0 \\
0 & 0.2^2
\end{bmatrix}\right). 
\end{equation}
We consider NNME with the following models on $X$: (a) True distribution of $X$ (benchmark). (b) A correct parametric model, i.e., assuming a 2-component Gaussian-mixture with unknown parameters. (c) A misspecified parametric model, e.g., a 4-component Gaussian-mixture or (after centering and scaling) a standard bivariate $t$-distribution. (d) The NICE model. The performance of every version is evaluated by the ISE, calculated on a uniform grid on $[-1, 0.2] \times [-1, 0.5]\cup[-0.5, 1] \times [-0.2, 1]$. The results are shown in Figure~\ref{fig:learnX}.}

We consider settings where the sample size $n$ varies in $\{1000,4000,8000\}$ and the standard deviation $\sigma_0$ of measurement errors varies in $\{0.05, 0.2\}$. Fixing a model for $X$, we also vary the structures of the two neural networks for representing $f$ (decoder) and approximating the posterior distribution of $X$ (encoder). Suppose the encoder has $\ell_1$ layers and the decoder has $\ell_2$ layers, with 32 nodes per layer; we consider four values of $(\ell_1,\ell_2)$, $\{(6,3)$, $(6,5)$, $(9,3)$, $(9,5)\}$, corresponding to the 4 box plots associated with each method in each panel of Figure~\ref{fig:learnX}. 

When the measurement error is small (e.g., top panels of Figure~\ref{fig:learnX}), the performance of NNME is insensitive to the chosen model for $X$. In this case, the posterior distributions of $x_i$'s are primarily determined by the observed $w_i$'s, and the prior is uninformative. When the measurement error is large (e.g., bottom panels of Figure~\ref{fig:learnX}), if the sample size is large (e.g., $n\in \{4000,8000\}$), then using a correct parametric model (i.e., a 2-component Gaussian mixture) yields similar performance as using the true model. The NICE model yields slightly worse performance, but it does not require specifying any parametric model. Regarding the two misspecified parametric models, the 4-component Gaussian mixture performs as well as (sometimes even better than) the true model, but the (unimodal) $t$-distribution performs unsatisfactorily. If the sample size is small (e.g., $n=1000$), then the performances of different models are again similar. The results suggest a practical guideline for choosing the model for $X$: We can use either the NICE model or a parametric model that allows for enough modes.

\section{Comparison with Classical Methods}
\label{sec:Comparison}

\subsection{A brief review} \label{subsec:literature}

Most classical methods for estimating MEMs start from a nonparametric regression method for the error-free case, and aim to address the bias caused by replacing $X$ with $W$ in the presence of measurement errors. Some notable approaches include the deconvolution approach \citep{Fan1993}, the regression spline approach \citep{Carroll1999,Jiang2018}, and the simulation extrapolation approach \citep{Carroll1999}.

The deconvolution approach builds on local smoothing estimators for nonparametric regression in the error-free case. A local smoother takes the form $\hat{f}(x)=\sum_{i=1}^n \ell_i(x) y_i$, where $\ell_i(x)$ is determined by $x$ and $x_1,x_2,\ldots,x_n$. Given a kernel function $K(\cdot)$ and a bandwidth $h$, the Nadaraya-Watson estimator corresponds to $\ell_i(x)=K(\frac{x_i-x}{h})/[\sum_{j=1}^n K(\frac{x_j-x}{h})]$. In the presence of measurement errors, $\{K(\frac{x_j-x}{h})\}_{j=1}^n$ are unobserved. \cite{Fan1993} proposed replacing $K(\frac{x_j-x}{h})$ by $L(\frac{w_j-x}{h})$, such that $\mathbb{E}[L(\frac{w_j-x}{h})|x_j]=K(\frac{x_j-x}{h})$. The function $L(\cdot)$ is defined by
\[
L(x) = \frac{1}{2\pi}\int e^{-\sqrt{-1}\, tx} \frac{\phi_K(t)}{\phi_\epsilon(t/h)}dt, \qquad
\left\{\begin{array}{l}
\phi_K(t): \mbox{Fourier transform of $K(\cdot)$},\\
 \phi_\epsilon(t): \mbox{characteristic function of $\epsilon$}.
 \end{array}\right.
\]
This method is inspired by a technique in density deconvolution problems \citep{carroll1988optimal,stefanski1990deconvolving} and, hence, called the deconvolution approach. This approach can be generalized to combine with other local smoothing estimators. For example, \cite{Delaigle2009} generalized the idea to the local polynomial estimator, where they proposed functions $\{L_m(\cdot)\}_{m\geq 0}$ such that $\mathbb{E}[(w_j-x)^mL_m(\frac{w_j-x}{h})|x_j]= (x_j-x)^m K(\frac{x_j-x}{h})$. 
In our numerical experiments, we include both the original deconvolution method (abbreviated as ``deconv'') and the generalization to local polynomials (abbreviated as ``lpoly''). See Table~\ref{tb:methods}.

One advantage of the deconvolution approach is that it requires no structural assumption on $f(\cdot)$, except the smoothness. Another advantage is that it attains the minimax rate of convergence \citep{Fan1993}. However, its practical performance crucially depends on the  bandwidth selection. For density estimation with error-in-variables, \cite{Delaigle2004,Delaigle2004a} proposed data-driven bandwidth selectors using cross-validation or bootstrap. There is relatively little work on bandwidth selection for nonparametric regression with error-in-variables. One exception is \cite{Delaigle2008}, where they combined cross-validation with the simulation extrapolation idea (to be introduced below).

The regression spline approach approximates $f(\cdot)$ by spline bases. Different from the error-free case, fitting splines with measurement errors requires either estimating the posterior distribution of $X$ or calculating an unbiased score function. \cite{Carroll1999} used the truncated power basis and replaced $x_i^m$ and $(x_i-\xi_j)_+^m$ by $\mathbb{E}(x_i^m|w_1,\ldots,w_n)$ and $\mathbb{E}[(x_i-\xi_j)_+^m|w_1,\ldots,w_n]$, where $\xi_j$'s are the fixed knots.  To compute these conditional moments, they assumed that the marginal distribution of $X$ is a mixture of normals and developed a Gibbs sampling algorithm to obtain the posterior distribution of $X$. \cite{Jiang2018} used the B-spline basis and proposed an estimating equation method. Denote by $\beta$ the spline coefficients and let $S_{\beta}^*(W,Y,\beta)$ be the gradient of the log likelihood of $(W,Y)$, assuming a working density on $X$. This method finds a function $a(X,\beta)$ such that $\mathbb{E}\{ S_{\beta}^*(W,Y,\beta)-\mathbb{E}^*[a(X,\beta)|W,Y] | X\}=0$, where $\mathbb{E}$ and $\mathbb{E}^*$ stand for the probability measure with $X$ following the true and working density, respectively. $\beta$ is estimated by the equation $\sum_{i=1}^n \{ S^*_\beta(w_i,y_i,\beta) - \mathbb{E}^*[a(X,\beta)|w_i,y_i]\}=0$. We abbreviate this method as ``BSSP''; see Table~\ref{tb:methods}. 

A key assumption made by the spline approach is that the density of $X$ has a compact support \citep{hall2005discrete}. Under this assumption, the rate of convergence can be faster than the minimax rate: For Gaussian measurement errors, when $f(\cdot)$ has a continuous $k$th derivative, the minimax rate for the mean squared error is as slow as $\log(n)^{-k}$ \citep{Fan1993}; but assuming a compact support for $X$, the spline approach can attain the standard nonparametric rate $n^{-\frac{2k}{2k+1}}$ \citep{Jiang2018}. The spline approach still requires choosing tuning parameters, which are the knots in spline construction. Following \cite{eilers1996flexible}, we use a large number of equally spaced knots and add a penalty on the difference of coefficients
 of adjacent splines. It reduces to choosing the penalization parameter $\lambda$. \cite{Carroll1999} introduced a double-smoothing technique for selecting $\lambda$.

The simulation extrapolation (SIMEX) approach \citep{cook1994simulation} builds upon an arbitrary nonparametric regression method for the error-free case. It then replaces $w_i$ by $w_i(\lambda) = w_i+\sigma_0\sqrt{\lambda}\delta_i$, where $\delta_i$ is a ${\cal N}(0,1)$ variable independent of data and $\sigma^2_0$ is the variance of measurement errors. Let $\hat{f}(x;\lambda)$ be the estimated regression function by plugging in $\{w_i(\lambda)\}_{1\leq i\leq n}$ as covariates. In a parametric MEM, $\hat{f}(x;\lambda)$, as a curve of $\lambda$, gives a consistent estimator of the true $f(x)$ when extrapolated to $\lambda=-1$ \citep{cook1994simulation}. This idea was generalized by \cite{Carroll1999} to nonparametric MEMs: It first computes $\hat{f}(x;\lambda)$ for a few values of $\lambda\geq 0$ using some nonparametric regression method, and then extrapolates to $\lambda = -1$.  The extrapolation is often done by fitting a quadratic curve of $\lambda$. 

SIMEX provides a convenient way to extend arbitrary nonparametric regression methods from the error-free case to the error-in-variable case. However, its consistency requires that the variance of measurement errors tends to zero. It was argued in \cite{cook1994simulation} that in a parametric MEM the simulation extrapolation with a quadratic extrapolant has an asymptotic bias of $O(\sigma_0^6)$; hence, it is consistent only if $\sigma_0\to0$ as $n\to\infty$. The use  of SIMEX cannot avoid tuning parameter selection in the original nonparametric regression method. When the nonparametric regression method in SIMEX is the local polynomial estimator,  \cite{staudenmayer2004} introduced the empirical bias band width selector (EBBS) for bandwidth selection. EBBS estimates the bias of SIMEX for each fixed $h$ and selects $h$ that minimizes the mean-squared error.

All the above classical methods for fitting MEMs are primarily designed for dimension $d=1$. When $d>1$, although these methods have natural extensions, their implementations are much more challenging: the construction of kernels or splines, the computation of estimators, and the selection of bandwidth or knots, all become substantially more difficult. In our numerical experiments, we find it quite difficult to implement these methods when $d>1$; in particular, the computation of some methods is expensive even for a moderately large $n$. In contrast, our proposed neural network approach is a promising alternative, and can be cheaply implemented for a wide rage of $(d,n)$. 

In numerical experiments we also include the kriging methods for Gaussian process regression as a competitor. Different from nonparametric regression, Gaussian process regression assumes that $f(x)$ is randomly generated from a stationary Gaussian process (SGP). It is common to use a radial basis exponential kernel in the SGP, i.e.,  
\begin{equation} \label{SGP}
Y=f(X)+\epsilon, \ \ f(x)\sim \mathrm{SGP}(0, K(\cdot,\cdot)),\ \ K(x,y)=\tau^2e^{-\beta\|x-y\|^2},\ \
\epsilon\sim {\cal N}(0,\sigma^2 I_n). 
\end{equation}
Without measurement error, the best linear unbiased predictor of $f(x)$ at any given $x$ is $\hat{f}(x) = K(x, {\bf x}_n)(K({\bf x}_n, {\bf x}_n) + \sigma^2I_n)^{-1}{\bf y}_n$, where ${\bf y}_n=\{y_i\}_{1\leq i\leq n}$, $K(x,{\bf x}_n)=\{K(x,x_i)\}_{1\leq i\leq n}$, and $K({\bf x}_n, {\bf x}_n)$ is a $n\times n$ matrix whose $(i,j)$th entry is $K(x_i,x_j)$. The kriging method first estimates $(\sigma^2, \tau^2,\beta)$ by maximizing the likelihood and then plugs these parameters into the best linear unbiased predictor. In the presence of measurement errors, \cite{Cressie2006} proposed a variant of the best linear unbiased predictor by replacing $K(x, {\bf x}_n)$ by $\tilde{K}_*(x,{\bf w}_n)$ and $K({\bf x}_n, {\bf x}_n)$ by $\tilde{K}({\bf x}_n, {\bf x}_n)$, where 
\[
  \tilde{K}(w_i, w_j) =\begin{cases} \tau^2 \frac{\exp\bigl(-\frac{\beta}{1 + 4\beta\sigma^2_0}\|w_i - w_j\|^2\bigr)}{(1 + 4\beta\sigma^2_0)^{d/2}},& w_i\neq w_j, \cr
  \tau^2, & w_i=w_j
  \end{cases}
\qquad \tilde{K}_*(x, w_i) = \tau^2 \frac{\exp(-\tfrac{\beta}{1 + \beta\sigma^2_0}\|w_i - w_j\|^2)}{(1 + \beta\sigma^2_0)^{d/2}}. 
\]
Same as before, $\sigma_0^2$ is the variance of measurement errors. The estimator of $f(x)$ is a plug-in version where $(\tau, \beta, \sigma)$ are estimated by maximizing a pseudo-likelihood. Although this method is not designed for nonparametric regression, we include it as a competitor of the neural network approach, especially for $d>1$.
%, since it is hard to implement the classical nonparametric/semi-parametric methods in the case of $d>1$. 
In our numerical experiments, we include both this kriging method that accounts for measurement errors (abbreviated as ``KALE''), as well as the standard kriging for the error-free case (abbreviated as ``KILE'').

\begin{table}[!hbt]
\centering
\caption{Methods included in numerical experiments.} \label{tb:methods}
\scalebox{.96}{\begin{tabular}{c | l  | c | l  } %|c|c|c|c|c|c|c|
 \toprule
 Method & Description  &  Method & Description \\
 \hline
deconv & deconvolution  
& SIMEX & simulation extrapolation\\
lpoly* & local polynomial regression ($*$) & KILE & kriging with a radial basis kernel ($*,\dag$) \\
lpoly & local polynomial deconvolution  & 
KALE & kriging, accounting for MEM ($\dag$)\\
Pspline & penalized regression splines ($*$)  &
NN & neural networks for regression ($*$) \\
BSSP & B-splines for MEM & 
NNME & neural networks for MEM \\
 \bottomrule
\end{tabular}}
$*$: ignoring measurement errors. $\dag$: assuming {\it a priori} that $f$ follows a Gaussian random field.
\end{table}

\subsection{Example 1: Deterministic functions} \label{subsec:Simu1}
We consider a  1-dimensional smooth regression function
\[
f(x) = \frac{\sin((3x-1.5)\pi)}{1 + 4(6x-3)^2[\mathrm{sgn}(2x-1)+1]}, 
\]
which was used in the simulation study of \cite{berry2002bayesian}. We generate data as in model \eqref{eqn:model}, where $U\sim {\cal N}(0,\sigma_0^2)$, $\epsilon\sim {\cal N}(0,\sigma^2)$, and $X$ is generated from either a uniform distribution or a beta distribution.

We implemented all the methods in Table~\ref{tb:methods}. Since the focus of this paper is the neural network approach, we prefer not to investigate tuning parameter selection for other methods in detail. To this end, we use the {\it ideal tuning parameters} for most classical methods, defined as the tuning parameters that minimize the true loss function (e.g., ISE), which biases slightly in favor of these classical methods. 
%On the contrary, we use data-driven tuning parameters for neural network approaches, which are selected by cross-validation as described in  Section~\ref{subsec:NNME}.
In Pspline and BSSP, we use cubic B-splines with equally spaced knots. We run the methods for different numbers of knots, from 5 to 30,  and choose the one that minimizes the true loss function. Similarly, we run the methods for a range of values of the penalization parameter and select the ideal one. BSSP requires initial estimates of spline coefficients. We initialize by fitting B-splines on the observed data, ignoring measurement errors. In SIMEX, the original nonparametric regression method is the regression spline with truncated power bases. The selection of knots and penalization parameter, as well as initialization of coefficients, is similar to that for BSSP. In ``deconv,'' we set the bandwidth as $\sigma_0 [\log(n)]^{-1/2}$, which is the theoretically optimal one for density deconvolution under Gaussian measurement errors \citep{stefanski1990deconvolving}. In ``poly*'' and ``poly,'' we set the polynomial degree as 1. We run the methods for 5 bandwidth values near $1.06\sigma_0n^{-1/5}$ and select the one that minimizes the true loss function. 

The tuning parameter selection for NNME has been described in Section~\ref{subsec:NNME}. In this simulation, we fix the values of tuning parameters, instead of using data-driven ones. We use 6 hidden layers in the decoder and 6 hidden layers in the encoder, where each layer has 32 nodes. We use a parametric model for the distribution of $X$, which is $2\cdot t_3$ (after centering and scaling),  fix the L2-regularization parameters $\lambda_0$ and $\lambda_1$ as $10^{-5}$, and set $\lambda_2=0$.

Recall that $\sigma^2$ and $\sigma^2_0$ are the conditional variances of $Y\mid X$ and $X\mid W$, respectively. We consider two scenarios: (a) Large response error and small measurement error, where $\sigma = 0.3$ and $\sigma_0 = 0.1$; and (b) small response error and large measurement error, where $\sigma = 0.1$ and $\sigma_0 = 0.2$. 
For each scenario, we generate data from two distributions of $X$, $\mathrm{Uniform}(0,1)$ and $\mathrm{Beta}(2,2)$, where in the latter case fewer $x_i$'s locate near boundaries. We let the sample size $n$ range in $\{500, 1000, 2000\}$. We evaluate the performance by 
the integrated squared error (ISE), 
\[
\int_0^1 [\hat{f}(x)-f(x)]^2dx,
\]
which is computed by 1000 points equally spaced in $[0,1]$. The results for scenarios (a) and  (b) are shown in Tables~\ref{tab:EX1-smallErr} and Table~\ref{tab:EX1-largeErr}, respectively.

\begin{table}[!tb]
\centering
\caption{Example 1 (smooth function), small measurement error ($\sigma_0 = 0.1$, $\sigma = 0.3$). The ISE and its standard deviation (in brackets) are based on 50 repetitions.} \label{tab:EX1-smallErr}
\scalebox{.95}{\begin{tabular}{ ccccccc} %|c|c|c|c|c|c|c|
 \toprule
  &  \multicolumn{3}{c}{$X\sim \mathrm{Uniform}(0,1)$}&  \multicolumn{3}{c}{$X\sim \mathrm{Beta}(2,2)$}\\ 
 \cmidrule(r){2-4}   \cmidrule(r){5-7}
& n=500 & n=1000 & n=2000 & n=500 & n=1000 & n=2000 \\
  \midrule
 BSSP & .026 (.003) & .020 (.003) & .013 (.002) & .055 (.005) & .051 (.005) & .045(.004) \\ 
Pspline & .050 (.002) & .048 (.001) & .047 (.01) & .089 (.003) & .089 (.002) & .085 (.002) \\ 
SIMEX & .015 (.001) & .009 (.001) & .006 (.000) & 0.025 (.002) & .020 (.002) & .016 (.001) \\ 
 deconv & .033 (.002) & .027 (.001) & .021 (.001) & .055 (.005) & .045 (.005) & .036 (.004)\\ 
lpoly & .039 (.002) & .034 (.001) & .029 (.001) & .077 (.003) & .077 (.003) & .071(.002)\\ 
 KALE & .052 (0.012) & .098 (.017) & .250(.012) & .076 (.006) & .078 (.006) & .102 (.009) \\
 KILE &  .050 (.002) & .048 (.001) & .046 (.001) & .086 (.002) & .085 (.002) & .082 (.002)\\ 
 NN & .047 (.002) &   .045 (.001) &    .045 (.001) & .092 (.003)& .081(.002) &   .082(0.002)\\ 
 NNME &  \textbf{.013} (.001)  &  \textbf{.008} (.001)  & \textbf{.005} (.000) & \textbf{.024} (.004)  &  \textbf{.015} (.003)  &  \textbf{.011} (.001)\\ 
%\midrule
% BSSP & .240 (.002) & .243 (.002) & .243 (.003) & .248 (.003) &  .250 (.003) &  .250 (.003)  \\ 
%Pspline & \textbf{.237} (.002) & .239 (.002) & .240 (.003) & .255 (.003) & .258 (.003) &  .259 (.003) \\ 
%lpoly* & .239 (.002) & .240 (.002) & .241 (.003) & .254 (.003) & .256 (.002) & .257 (.003)\\
% KALE & .258 (.006)  & .286 (.009)  & .362 (.007)  & .267 (.004) & .270 (.004) & .281 (.005) \\
% KILE &  .238 (.002)  & .239 (.002)  & .241(.003)  & .256 (.003) & .258 (.003) & .259 (.003)\\ 
% NN & .239 (.002)  &    .239 (.002)  &    \textbf{.236} (.002) & .258 (.003) &   .252 (.003) &    .251 (.003)\\ 
% NNME &  .239 (.002)  & \textbf{.238} (.003)  &    .237 (.003) & \textbf{.239} (.002)  &  \textbf{.235} (.003)   &  \textbf{.235} (.002) \\ 
 \bottomrule
\end{tabular}}
%\raggedright\hspace{0.8cm} Each entry shows mean (standard error) over 50 repetitions. 
\end{table}

\begin{table}[!tb]
\centering
\caption{Example 1 (smooth function), large measurement error ($\sigma_0 = 0.2$, $\sigma = 0.1$). The ISE and its standard deviation (in brackets) are based on 50 repetitions.} \label{tab:EX1-largeErr}
\scalebox{.95}{\begin{tabular}{ ccccccc} %|c|c|c|c|c|c|c|
  \toprule
  &  \multicolumn{3}{c}{$X\sim \mathrm{Uniform}(0,1)$}&  \multicolumn{3}{c}{$X\sim \mathrm{Beta}(2,2)$}\\ 
 \cmidrule(r){2-4}   \cmidrule(r){5-7}
& n=500 & n=1000 & n=2000 & n=500 & n=1000 & n=2000 \\
  \midrule
 BSSP &  .164 (.007) &  .170 (.005) & .168 (.005) & .279 (.007) & .270 (.005) & .259 (.006)\\ 
Pspline & .185 (.003) & .179 (.002) & .178 (.001) & .247 (.003) & .246 (.003) & .242 (.002)\\
SIMEX & .125 (.006) &  .112 (.004) & .111 (.002) & .215 (.006) & .208 (.005) & .197 (.004)\\
deconv & .140 (.006) &  .127 (.005) & .125 (.004) & .209 (.007) & .213 (.006) & .194 (.005) \\
 lpoly & .155 (.005) & .150 (.004) & .151 (.003) & .229 (.005) & .237 (.003) & .228 (.003)\\ 
 KALE &  .236 (.007) & .229 (.009) & .227 (.009) & .273 (.005) & .283 (.005) & .282 (.005) \\ 
 KILE &  .185 (.003) & .179 (.002) & .180 (.001) & .241 (.003) & .243 (.002) & .243 (.002)  \\
 NN &  .174 (.003) &    .173 (.003) &    .175 (.001)  & .239 (.005) & .244 (.003)&    .239 (.002)\\ 
 NNME & \textbf{.021} (.003) & \textbf{.019} (.002) & \textbf{.014} (.001) &  \textbf{.059} (.005) & \textbf{.058} (.004) &  \textbf{.045} (.003) \\
%\midrule
% BSSP &  .256 (.002) & .260 (.002) & .259 (.003) & .291 (.005) & .279 (.004) & .296 (.005)\\ 
%Pspline & .246 (.002) & \textbf{.245} (.002) & .247 (.002) & .275 (.003) & .273 (.003) & .279 (.002)\\
%lpoly* & \textbf{.245} (.002) & \textbf{.245} (.002) & .247 (.002) & .269 (.003) & .268 (0.003) & .274 (.002)\\
% KALE &  .270 (.004) & .275 (.004) & .282 (.004) & .281 (.004) & .284 (.003) & .287 (.003)\\ 
% KILE &  .246 (.002) & .246 (.002) & .247 (.002) & .275 (.003) &  .273 (.003) & .278  (.002)\\
% NN &  .251(.002) & .246 (.002) & \textbf{.241} (.002) & .274 (.003) &    .276 (.003) &   .266 (.003)\\ 
% NNME & .260 (.003) & .252 (.002) & .250 (.002) & \textbf{.244} (.002) & \textbf{.250} (.003) &   \textbf{.242} (.002)\\
  \bottomrule 
\end{tabular}}
\end{table}

%First, we look at the estimation performance, measured by ISE. 
When the measurement error is relatively small (Table~\ref{tab:EX1-smallErr}), NNME attains the smallest error in all settings. SIMEX is the second best method.  BSSP, deconv, and lploy perform similarly, among which BSSP is the best when $X\sim \mathrm{Uniform}(0,1)$ and deconv is the best when $X\sim\mathrm{Beta}(2,2)$. The two methods that ignore measurement errors, Pspline and NN, are significantly worse than their couterparts that account for measurement errors. The kriging methods, KILE and KALE, have unsatisfactory performance, especially for a large sample size. The reason is that these methods assume a very different model on $f(x)$. When the measurement error is large (Table~\ref{tab:EX1-largeErr}), NNME is significantly better than any other method. In all the settings, the error of NNME is only 10\%-30\% of the error of the second best method. It is worth noting that the performance of NN is comparable to other methods. It suggests that the advantage of NNME indeed comes from the careful neural network design to account for measurement errors, as elaborated in Section~\ref{sec:Method}. %Among the other methods, SIMEX and deconv perform better. 

\subsection{Example 2: Functions generated from a Gaussian process}
\label{subsec:Simu2}
\begin{figure}[!hbt]
\centering
\begin{subfigure}{0.325\textwidth}
\includegraphics[width=\textwidth, trim=0 40 0 30, clip=true]{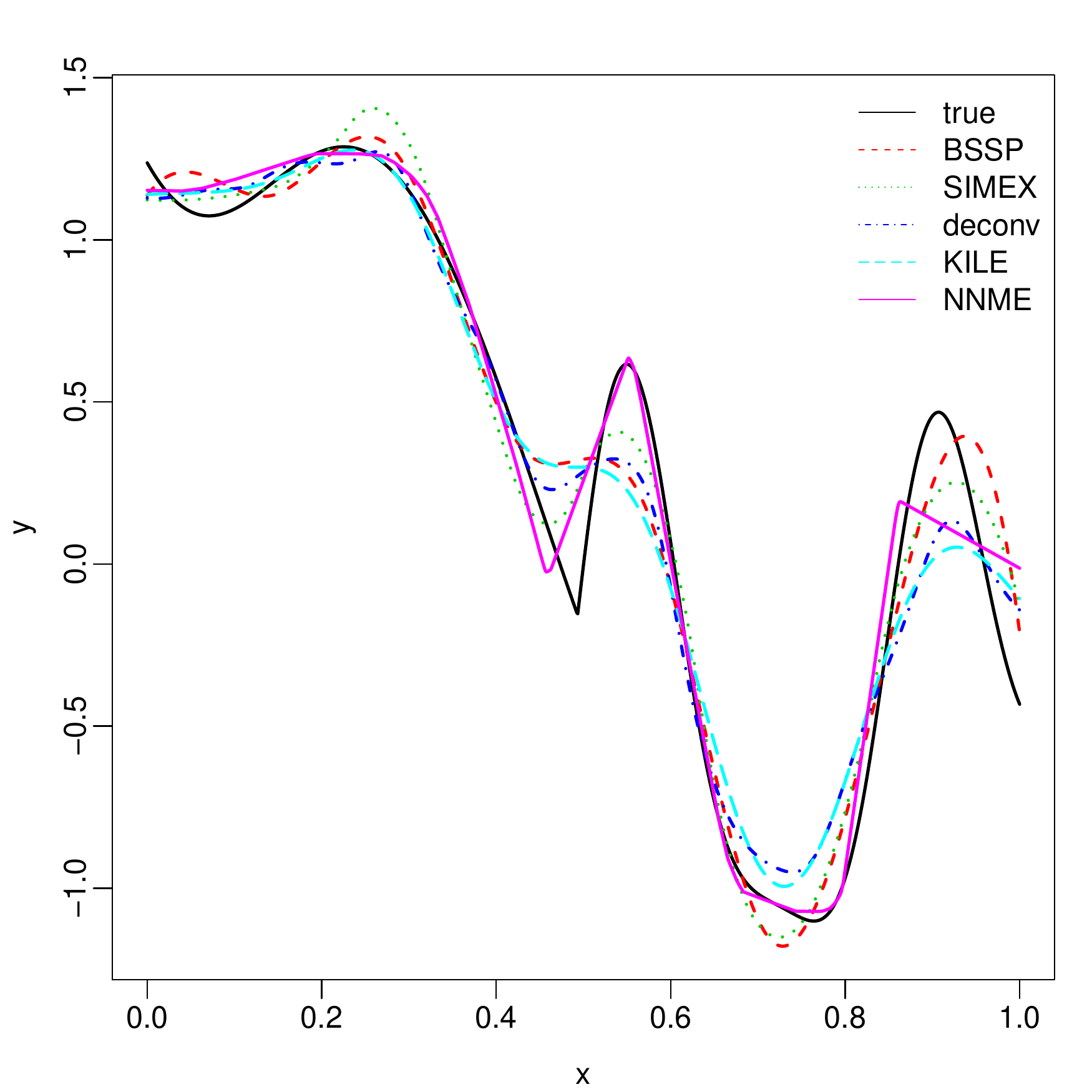}
\caption{$\sigma_0 = 0.05$, $\sigma = 0.2$, $n=500$.}
\end{subfigure}
\begin{subfigure}{0.325\textwidth}
\includegraphics[width=\textwidth, trim=0 40 0 30, clip=true]{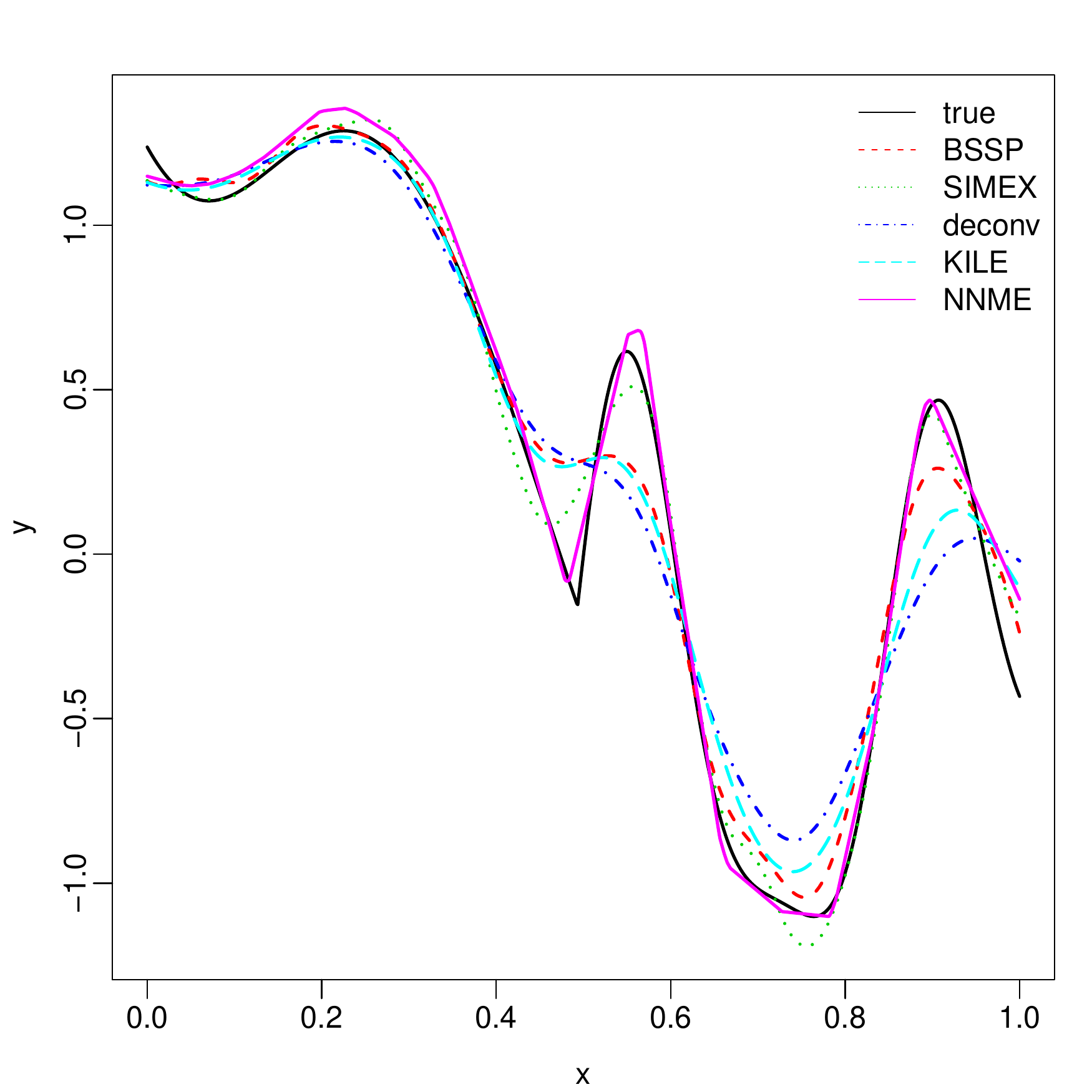}
\caption{$\sigma_0 = 0.05$, $\sigma = 0.2$, $n=2000$.}
\end{subfigure}
\begin{subfigure}{0.325\textwidth}
\includegraphics[width=\textwidth, trim=0 40 0 30, clip=true]{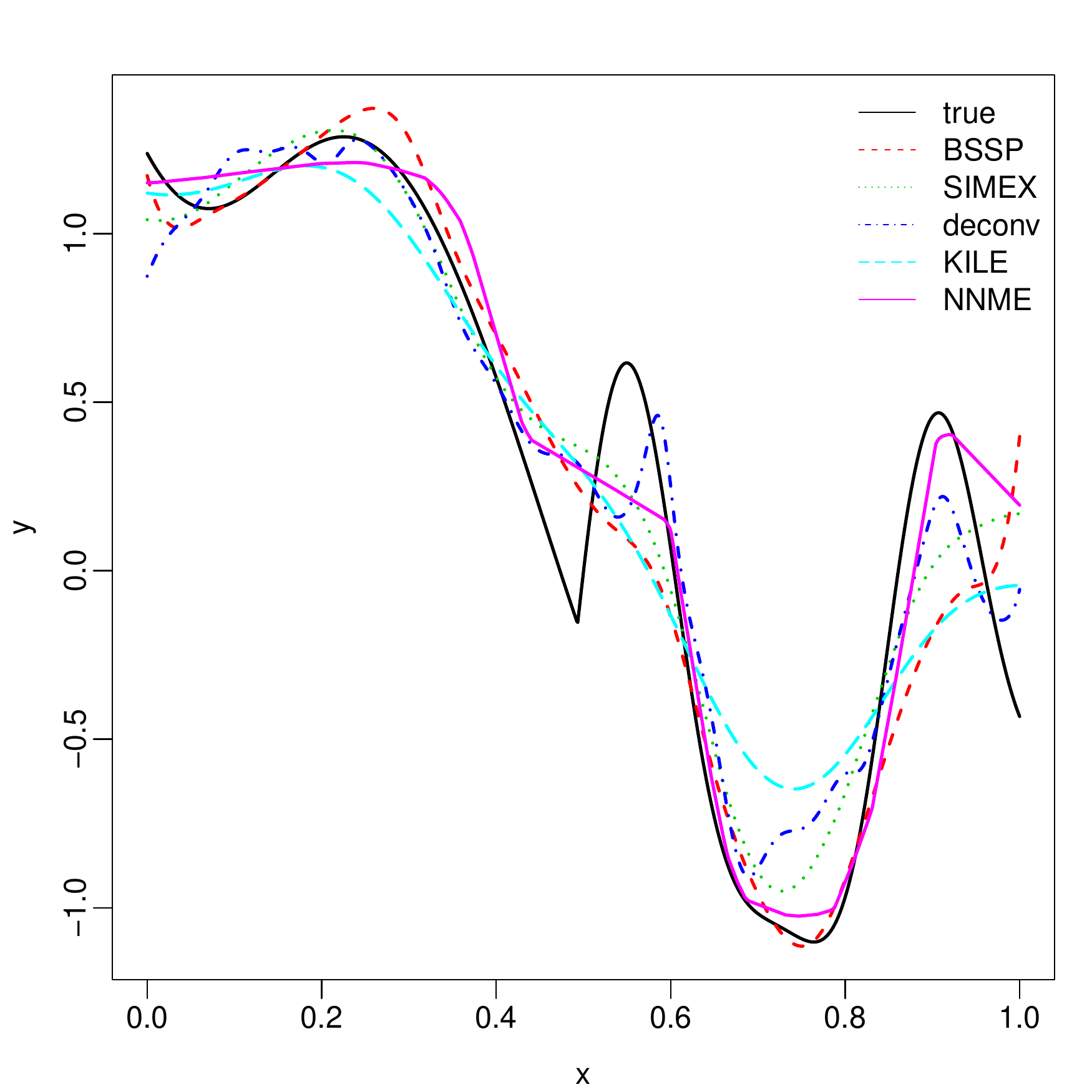}
\caption{$\sigma_0 = 0.5$, $\sigma = 0.1$, $n=2000$.}
\end{subfigure}
\caption{Example 2 (function generated from GP with a mixture of two kernels).}
\label{fig:GP_mix}
\end{figure}

We consider an example where $f(x)$ is generated from a Gaussian process (GP) on $[0,1]$ with two different radial basis function (RBF) kernels on $[0,0.5]$ and $[0.5,1]$, respectively. Both kernels are squared exponential kernels of the form 
\[
K(x,y)=\exp(-\beta|x-y|^2), \qquad \mbox{where}\quad \beta_1=16, \quad \beta_2=64.
\]
A GP with a squared exponential kernel has mean squared derivatives of all orders \citep{rasmussen2003gaussian} and is smooth with high probability. Therefore, $f(x)$ is smooth within $[0,0.5]$ and $[0.5,1]$, but not at $0.5$. In simulations, we generate $f(x)$ as follows: 
Without loss of generality, suppose $n$ is even and write $n=2m$. First, we sample $\{x_i\}_{i=1}^n$ uniformly from $[0,1]$ and relabel them so that $x_1<x_2<\ldots<x_n$. Next, we sample $\{f(x_i)\}_{1\leq i\leq m}$ from the first GP and $\{\tilde{f}(x_i)\}_{m\leq i\leq n}$ from the second GP.  Last, let $f(x_i)=\tilde{f}(x_i)+f(x_m)-\tilde{f}(x_{m})$ for $i=m,m+1,\ldots, n$, to make $f(x)$ continuous at $x_{m}$ (with high probability, $x_m$ is located near 0.5). Examples of $f(x)$ from simulations are shown in Figure~\ref{fig:GP_mix} by solid black curves.

Given $\{(x_i, f(x_i))\}_{i=1}^n$, we generate $\{(w_i,y_i)\}_{i=1}^n$ as in model \eqref{eqn:model}, where the measurement error follows a normal distribution ${\cal N}(0,\sigma_0^2)$. We fix $\sigma=0.2$ and let $\sigma_0$ take values in $\{0.02, 0.05, 0.1, 0.2\}$. For each choice of $\sigma_0$, we consider three different sample sizes, $n\in \{500, 1000, 2000\}$. The integrated squared errors (ISE) of different methods are reported in Table~\ref{tab:EX2}. Except for very small $\sigma_0$, NNME has the best performance, and SIMEX has the second best performance. KALE is significantly worse than the other methods; one possible reason is that the generating process of $f(x)$ is not the same as that in \eqref{SGP}.   The three methods that ignore measurement errors, NN, Pspline, and KILE, work well when the measurement error is small (e.g. $\sigma_0 = 0.02$), but their performance is significantly worse when the measurement error is large.

\begin{table}[!hbt]
\centering
\caption{Example 2 (function from Gaussian process). The ISE and its standard deviation (in brackets) is shown.} \label{tab:EX2}
\scalebox{.95}{\begin{tabular}{ ccccccc} %|c|c|c|c|c|c|c|
  \toprule
  &  \multicolumn{3}{c}{$\sigma_0=0.02$}&  \multicolumn{3}{c}{$\sigma_0=0.05$}\\ 
 \cmidrule(r){2-4}   \cmidrule(r){5-7}
& n=500 & n=1000 & n=2000 & n=500 & n=1000 & n=2000 \\
  \midrule
 BSSP & .015 (.003) & .012 (.002) & .011 (.002) & .035 (.006) & .033 (.005) & .031 (.005) \\
Pspline & .004 (.001) & .003 (.000) & .002 (.000) & .032 (.005) & .031 (.005) & .027 (.004) \\
SIMEX & {\bf.003} (.000) & {\bf.002} (.000) & {\bf.001} (.000) & {\bf.016} (.003) & .012 (.002) & .008 (.001) \\
deconv & .018 (.002) & .011(.002) & .007 (.001) & .028 (.005) & .022 (.004) & .016 (.003) \\
lpoly & .014 (.002) & .009 (.001) & .006 (.001) & .044 (.006) & .024 (.004) & .018 (.003) \\
KALE & .055 (.034) & .035 (.011) & .070 (.030) & .155 (.039) & .181 (.055) & .168 (.046) \\
KILE & .004 (.000) & .003 (.000) & .002 (.000) & .032 (.005) & .030 (.004) & .027 (.004) \\
NN & .015 (.006) & .009 (.003) & .006 (.002) & .039 (.016) & .037 (.009) & .036 (.006) \\
NNME & .006 (.001) & .003 (.000) & .002 (.000) & {\bf.016} (.003) & {\bf.009} (.002) & {\bf.006} (.001) \\
\midrule
  &  \multicolumn{3}{c}{$\sigma_0=0.1$}&  \multicolumn{3}{c}{$\sigma_0=0.2$}\\ 
 \cmidrule(r){2-4}   \cmidrule(r){5-7}
& n=500 & n=1000 & n=2000 & n=500 & n=1000 & n=2000 \\
  \midrule
BSSP & .103 (.017) & .092 (.017) & .092 (.016) & .328 (.036) & .328 (.039) & .319 (.042) \\
Pspline & .126 (.017) & .121 (.015) & .120 (.015) & .352 (.037) & .341 (.035) & .332 (.036) \\
SIMEX & .065 (.011) & .052 (.008) & .048 (.007) & .264 (.033) & .236 (.029) & .221 (.028) \\
deconv & .087 (.014) & .079 (.011) & .068 (.009) & .277 (.033) & .263 (.030) & .247 (.030) \\
lpoly & .094 (.015) & .089 (.013) & .081 (.011) & .301 (.034) & .295 (.031) & .285 (.031) \\
KALE & .193 (.028) & .297 (.068) & .341 (.060) & .359 (.039) & .397 (.056) & .402 (.056) \\
KILE & .136 (.018) & .124 (.015) & .122 (.015) & .353 (.039) & .341 (.036) & .334 (.036) \\
NN & .124 (.034) & .121 (.028) & .125 (.026) & .342 (.206) & .332 (.182) & .331 (.157) \\
NNME & {\bf.034} (.006) & {\bf.028} (.004) & {\bf.026} (.006) & {\bf.206} (.037) & {\bf.182} (.032) & {\bf.157} (.030) \\
 \bottomrule 
\end{tabular}}
\end{table}

We also plot the estimated curves from a few different methods in Figure~\ref{fig:GP_mix}. For such $f(x)$, there are two challenges of applying classical methods: First, estimating the first and second halves of the curves require different smoothing parameters, as the true function is more wiggly in the second half. Second, smoothing should be eliminated at around $x=x_{n/2}$, as the true function is non-smooth at that point. These challenges indeed affect the performance of classical methods. On panel (b) of Figure~\ref{fig:GP_mix}, the classical methods (deconv, BSSP, and SIMEX) fail to capture the local curvature around $x=x_{n/2}$ and oversmooth the second half of the curve. In contrast, NNME fits the true curve very well.  When the sample size reduces to 500 (see panel (a)), NNME fits the true curve well except at the boundary; in this case, there are fewer samples at the boundary, and NNME produces a nearly linear curve at the boundary as a result of using the ReLU activation functions. When $\sigma_0$ increases to 0.5 (see panel (c)), NNME fits well each half of the curve but misses the curvature around $x=x_{n/2}$; in this case, the measurement error is large, masking local curvature of the true curve. In all three settings, NNME still attains the smallest ISE among all methods (see Table~\ref{tab:EX2}). 

We note that the performance of classical methods (deconv, BSSP, and SIMEX) can be improved by using different bandwidths or different sets of knots in $[0, x_{n/2}]$ and $[x_{n/2}, 1]$, respectively, but this more or less requires some prior knowledge. The neural network approach can adaptively impose different levels of smoothing in two regions, without any prior knowledge.

\subsection{Example 3: Two-dimensional functions} \label{subsec:Simu3}

We consider settings with $d=2$. It is hard to implement the three classical methods (deconv, BSSP, SIMEX), as the codes we downloaded were designed only for $d=1$. Therefore, we only compare NNME with two kriging methods (KILE and KALE). Recall that KILE ignores measurement errors and KALE assumes Gaussian measurement errors. 

\begin{figure}[!hbt]
\centering
\begin{tabular}{cc}
\begin{subfigure}{0.4\textwidth}
\includegraphics[width=\textwidth, trim=0 0 0 30, clip=true]{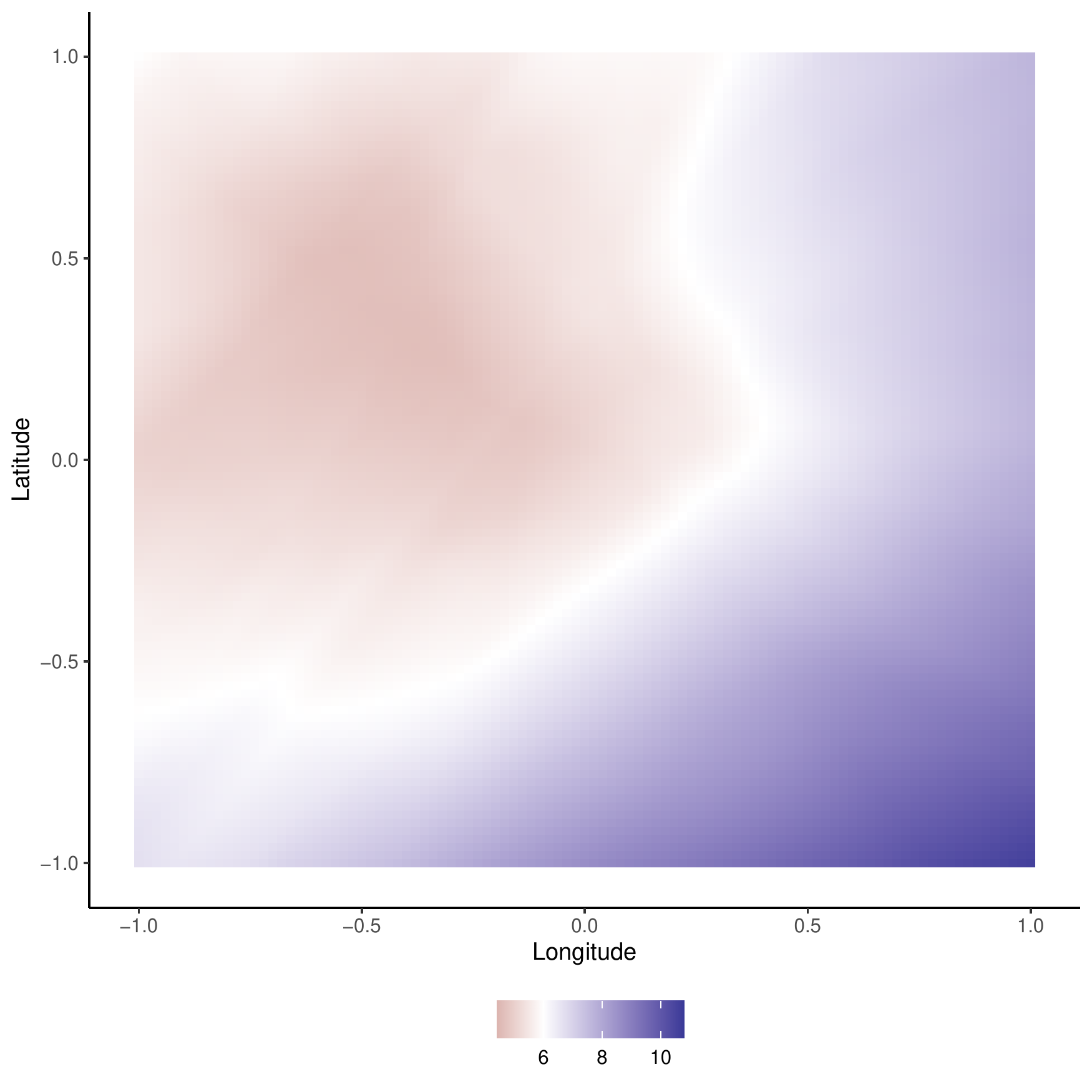}
\caption{True function.}
\end{subfigure}
 &
 \begin{tabular}{c}
\begin{subfigure}{0.52\textwidth}
\includegraphics[width=\textwidth, trim=0 0 0 25, clip=true]{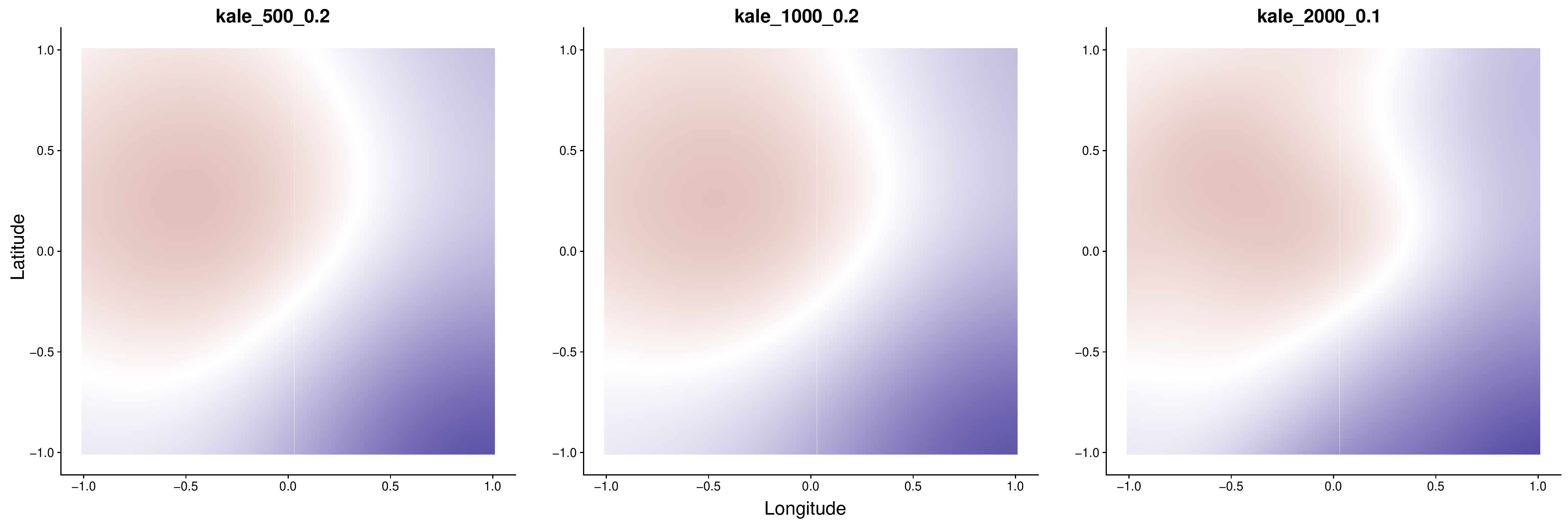}
\caption{KALE estimates with different $(n,\sigma_0)$.}
\end{subfigure} \\

\begin{subfigure}{0.52\textwidth}
\includegraphics[width=\textwidth, trim=0 0 0 25, clip=true]{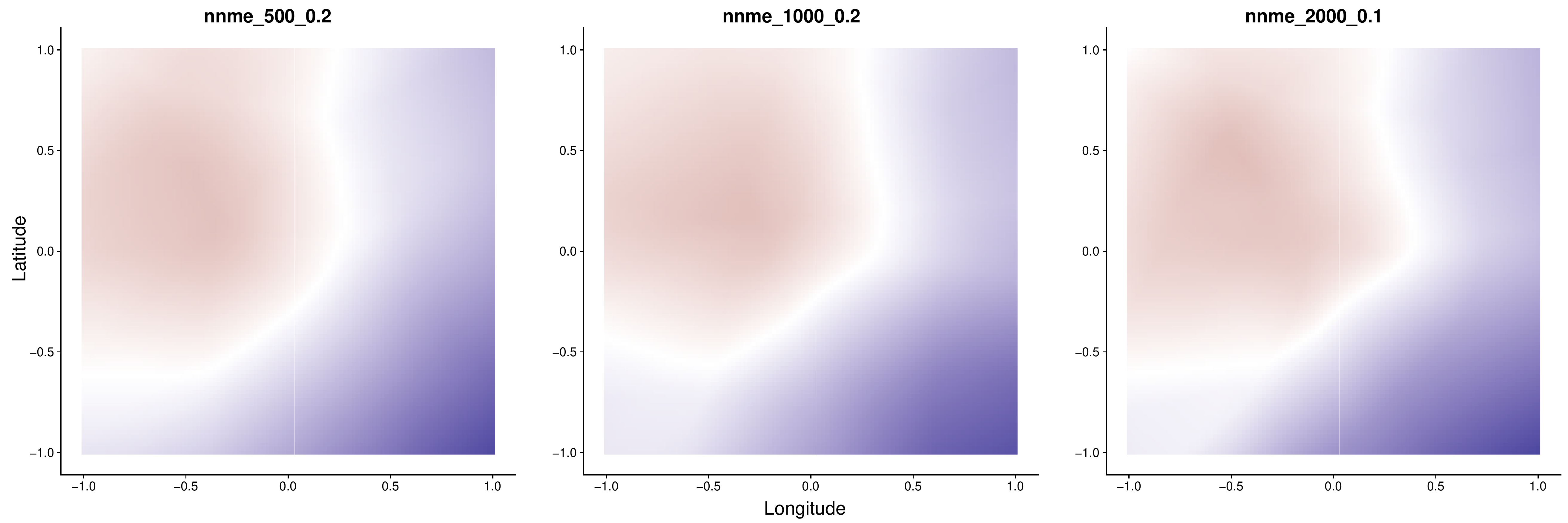}
\caption{NNME estimates with different $(n,\sigma_0)$.}
\end{subfigure}
\end{tabular}
\end{tabular}
\caption{Example 3-1 (two-dimensional function generated from a neural network). In (b)-(c), the three panels from left to right correspond to $(n,\sigma_0)=(500, 0.2)$, $(1000, 0.2)$, and $(2000, 0.1)$
%In (b)-(c), the title ``A\_X\_Y'' indicates the simulation setting, where A is the method, X is the sample size and Y is the measurement error. 
}
\label{fig:nn_pred}
\end{figure}

\paragraph{Example 3-1: Two-dimensional functions generated from a neural network.} We let $f: [-1,1]^2\to \mathbb{R}$ be defined by a neural network with 5 fully connected layers, 32 nodes per layer, and ReLU as activation function. The weights and thresholds in the neural network are randomly sampled from ${\cal N}(0,1)$ in the first layer and ${\cal N}(0, 0.2^2)$ in the rest of layers. One realization of $f(x)$ is shown in Figure~\ref{fig:nn_pred} (a).

\begin{table}[!hbt]
\caption{Example 3-1 (two-dimensional function from a neural network). The ISE evaluated at a $72\times72$ uniform grid, as well as its standard deviation (in brackets), is shown.} \label{tab:EX3-nn}
\scalebox{.9}{
\begin{tabular}{ ccccccc} %|c|c|c|c|c|c|c|
 \toprule
 &  \multicolumn{3}{c}{$\sigma_0 = 0.1,\ \sigma = 0.2$}&  \multicolumn{3}{c}{$\sigma_0 = 0.2,\ \sigma = 0.2$}\\
 \cmidrule(r){2-4}
  \cmidrule(r){5-7}
   & n=500 & n=1000 & n=2000 & n=500 & n=1000 & n=2000\\
\midrule
KILE &  .0071 (.0006)& .0052 (.0005)& .0041 (.0004)& .0170 (.0020)& .0148 (.0017)& .0133 (.0015) \\ 
 KALE &  \textbf{.0068} (.0005)& .0049 (.0004)& .0037 (.0004)& .0143 (.0017)& .0115 (.0012)& .0096 (.0010)  \\
 NN &  .0079 (.0007)& .0051 (.0004)& .0042 (.0004)& .0187 (.0024)&  .0145 (.0018)& .0136 (.0018)\\ 
 NNME & .0070 (.0003) & \textbf{.0046} (.0003) & \textbf{.0034} (.0003) & \textbf{.0135} (.0010) & \textbf{.0091} (.0008) & \textbf{.0074} (.0008) \\
  \bottomrule 
\end{tabular}}
%\raggedright\hspace{0.8cm} Each entry shows mean (standard error) over 50 repetitions.
\end{table}

Given $f(x)$, we generate $\{x_i\}_{i=1}^n$ uniformly from $[-1,1]^2$. Next, we generate $w_i$'s and $y_i$'s according to model \eqref{eqn:model}, where the measurement error has a multivariate normal distribution ${\cal N}(0, \sigma_0^2 I_2)$. We fix $\sigma=0.2$ and let $\sigma_0$ takes values in $\{0.1, 0.2\}$. We also vary the sample size by letting $n$ range in $\{500, 1000, 2000\}$. For NNME, we let the decoder network have 5 layers and the encoder network have 2 layers, where each layer has 32 nodes; we use a parametric model for $X$, assuming that all the coordinates of $X$ are independently distributed as $2\cdot t_3$. NN has only a decoder network, which also consists of 5 layers with 32 nodes per layer. 
We measure the performance of each method by the ISE, evaluated on a $72\times 72$ uniform grid. The ISE averaged over 10 repetitions is reported in Table~\ref{tab:EX3-nn}. It suggests that NNME has the best or nearly the best performances in all settings. The two methods ignoring measurement errors, NN and KILE, underperform their respective counterpart. KALE has a similar performance as NNME when $\sigma_0=0.1$, but its performance is inferior to NNME's when $\sigma_0=0.2$.

\begin{figure}[!hbt]
\begin{tabular}{cc}
\begin{subfigure}{.35\textwidth}
\includegraphics[width=\textwidth, trim=20 0 10 40, clip=true]{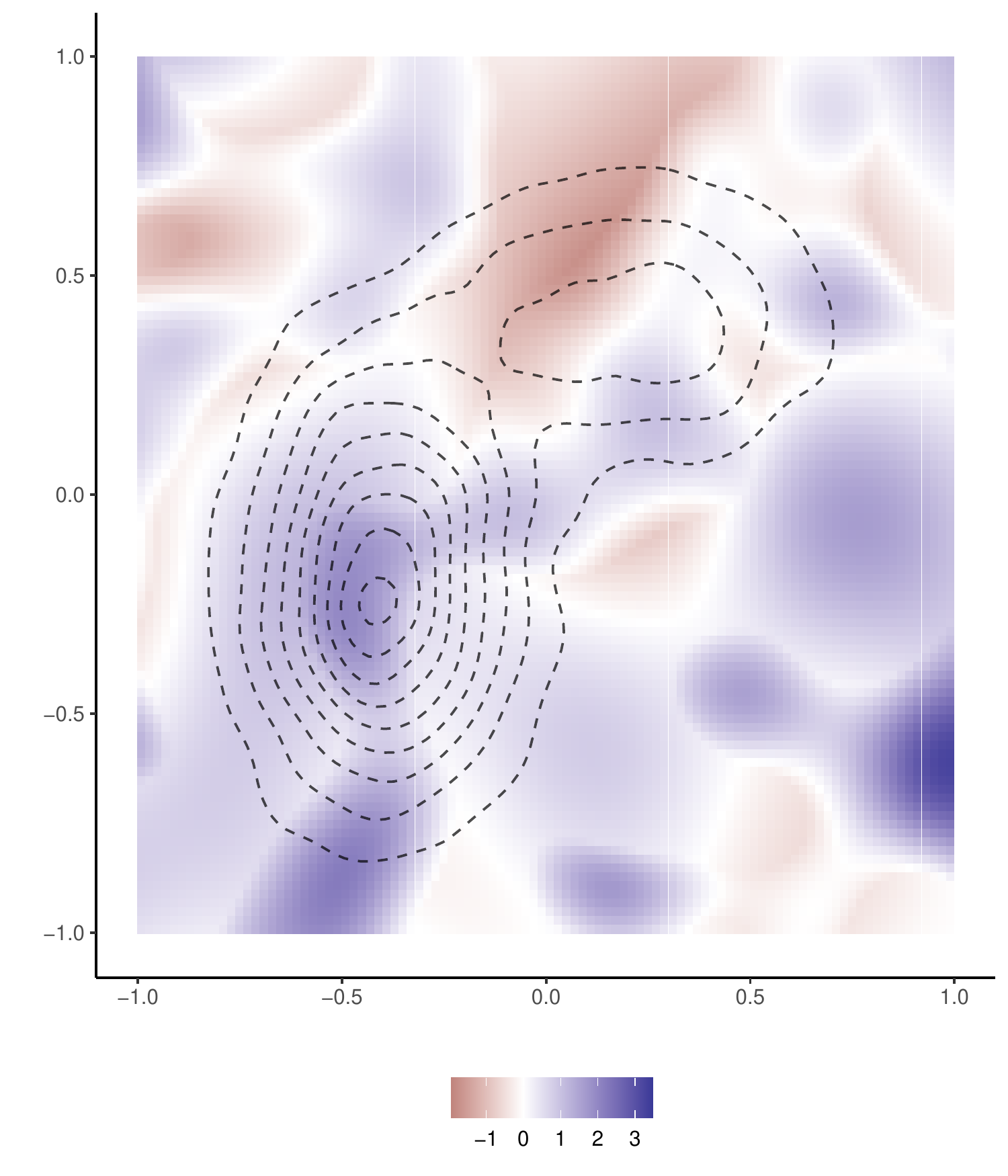}
\caption{True function.}
\end{subfigure}
&
\begin{tabular}{c}
\begin{subfigure}{.45\textwidth}
\includegraphics[width=\textwidth, trim=20 20 10 40, clip=true]{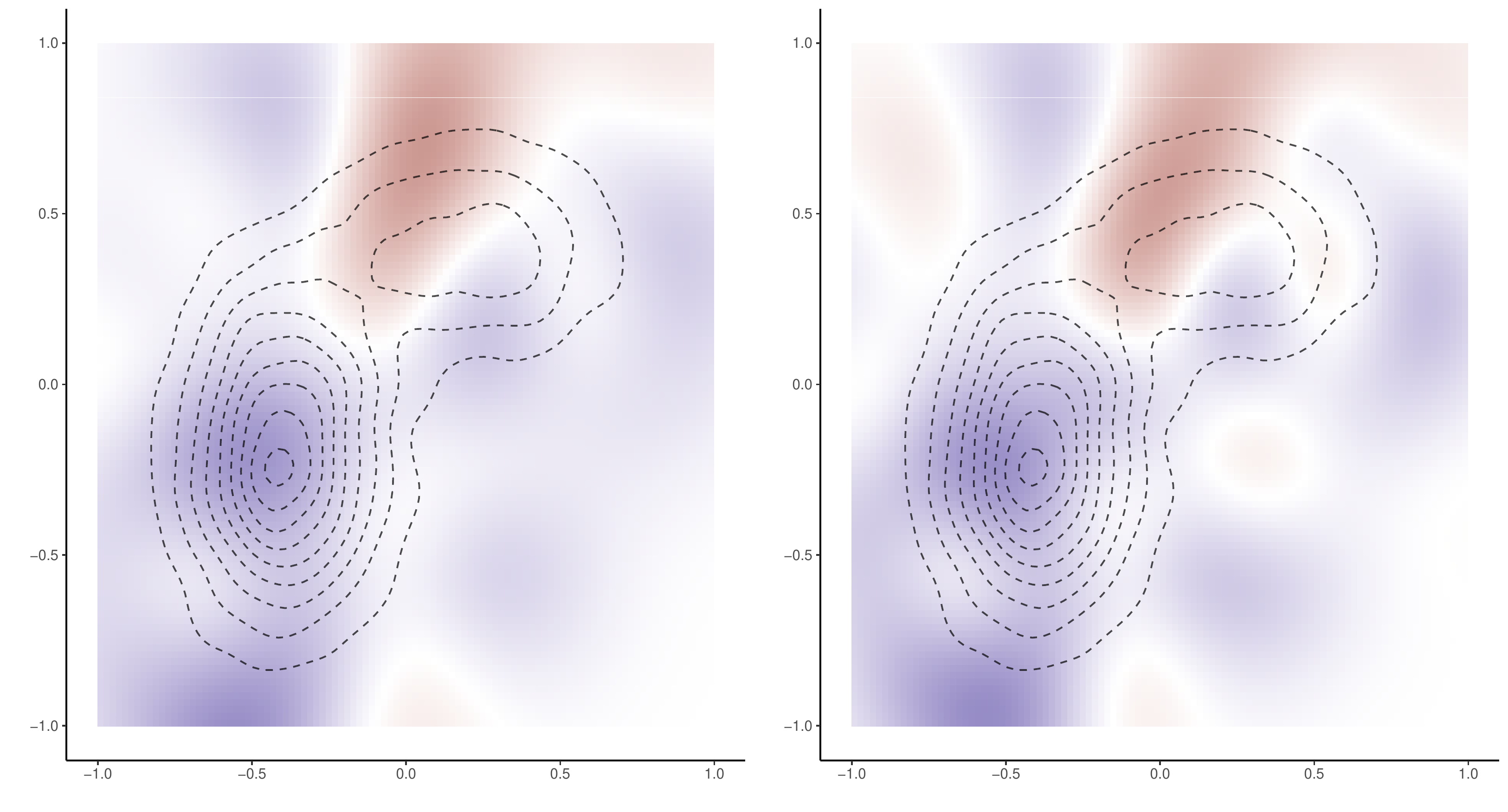}
\caption{KALE   ($n=500,2000$).}
\end{subfigure} \\
\begin{subfigure}{.45\textwidth}
\includegraphics[width=\textwidth, trim=20 20 10 40, clip=true]{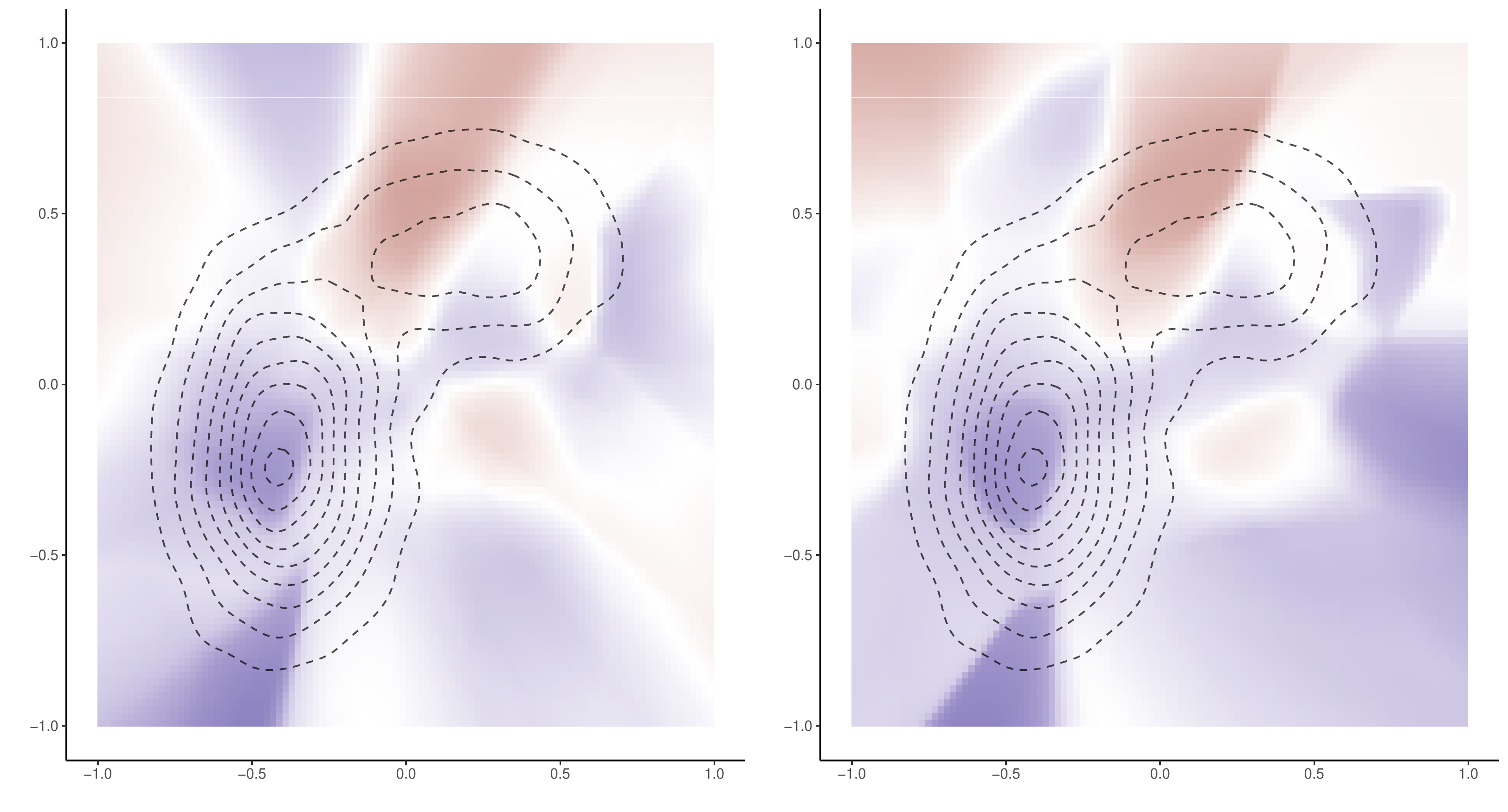}
\caption{NNME\_GM4  ($n=2000,8000$).}
\end{subfigure} \\
\begin{subfigure}{.45\textwidth}
\includegraphics[width=\textwidth, trim=20 20 10 40, clip=true]{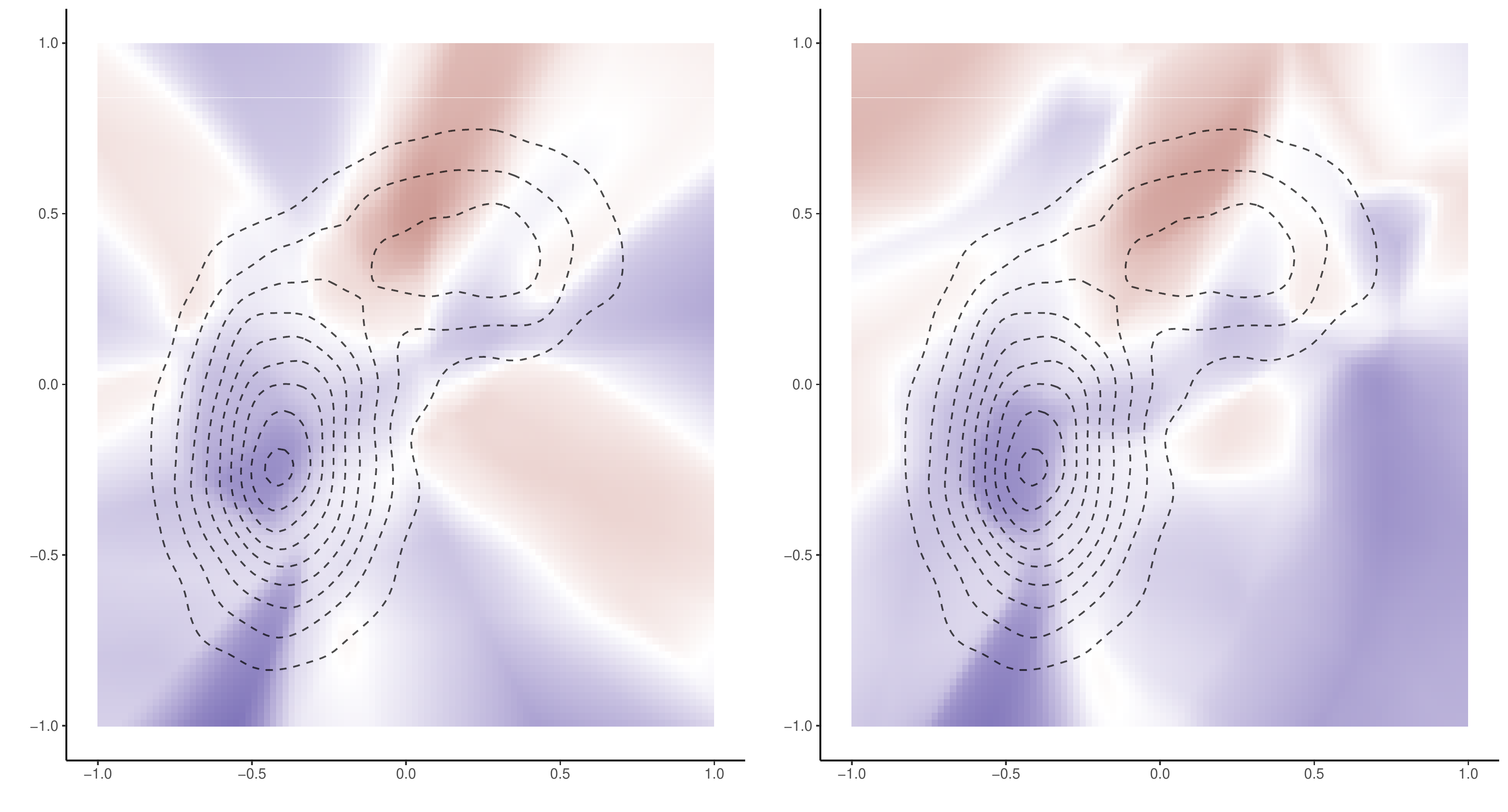}
\caption{NNME\_NICE  ($n=2000,8000$).}
\end{subfigure}
\end{tabular}
\end{tabular}
\caption{Example 3-2 (two-dimensional functions generated from a Gaussian process), with measurement error $\sigma_0=0.1$, and their estimates. The dashed lines are contours of the density of the training samples.}
\label{fig:prediction}
\end{figure}

\paragraph{Example 3-2: Two-dimensional functions generated from Gaussian processes.}
To generate $f$ from a 2-dimensional Gaussian processes as in \eqref{SGP}, we first generate $\{x_i\}_{1\leq i\leq n}$ from the distribution of $X$ and construct an $n\times n$ covariance matrix $\Sigma$, with $\Sigma_{ij}=\exp(-\beta\|x_i-x_j\|^2)$, and then generate $(f(x_1),f(x_2),\ldots,f(x_n))$ from ${\cal N}(0, \Sigma)$. We consider two settings. In the first one, we generate $f(x)$ with $\beta=16$. In the second one, we generate $f_1(x)$ and $f_2(x)$ with $\beta=16$ and $\beta=4$, respectively, and let $f(x)=\max\{f_1(x), f_2(x)\}$. 
An example of the realized $f(x)$ from the first setting is shown in Figure~\ref{fig:prediction}.

For each setting, we set the distribution of  $X$ to be a 2-component mixture of multivariate normal distributions, $0.7{\cal N}(\mu_1,\Sigma_1)+0.3{\cal N}(\mu_2, \Sigma_2)$, where $(\mu_1,\Sigma_1)$ and $(\mu_2,\Sigma_2)$ are the same as in \eqref{simu-X-distribution}. 
Given the $x_i$ and $f(x_i)$, we generate the $w_i$ and $y_i$ by adding Gaussian errors in the same way as in Example 3-1. We fix $\sigma=0.2$,  
and consider different values of $(n,\sigma_0)$, where $\sigma_0\in \{0.1, 0.2\}$ and $n\in \{500, 1000, 2000, 4000, 8000\}$. The performance metric ISE is computed via a uniform grid on the union of two rectangular regions, $[-1, 0.2] \times [-1, 0.5]\cup[-0.5, 1] \times [-0.2, 1]$. From the distribution of $X$, in the area outside these two rectangular regions, there are too few training $x_i$'s to learn the function $f$ correctly, regardless of which methods to use. We exclude such ``non-informative'' area, so that the ISE is informative for comparing differeing methods. 
In NNME, we use 10 layers in the network for representing $f$ and 6 layers in the network for approximating the posterior distribution of $X$, with 32 nodes per layer. How to choose the model for $X$ was discussed in Section~\ref{subsec:otherversion}, and the recommendation is to use either the NICE model or a mixture model that has a sufficient number of components. Here we consider both options: NNME\_NICE uses the NICE model, and NNME\_GM4 models $X$ by a 4-component  Gaussian mixture distribution with unknown parameters. We compare the two versions of NNME with the two kriging methods, KILE and KALE. The results are in Table~\ref{tab:EX3-gp}.

\begin{table}[!hbt]
\centering
\caption{Example 3-2 (two-dimensional function from Gaussian processes). The ISE evaluated at a uniform grid, as well as its standard deviation (in brackets), is shown. 
Top sub-table: $f$ is from a Gaussian process with $\beta=16$. Bottom sub-table: $f$ is the maximum of two Gaussian processes with $\beta=16$ and $\beta=4$, respectively. 
} \label{tab:EX3-gp}
\scalebox{.92}{\begin{tabular}{c cccccc} %|c|c|c|c|c|c|c|
  \toprule
 & & $n=$ 500 & 1000 & 2000 & 4000 & 8000\\
  \midrule
\multirow{4}{*}{$\sigma_0=0.1$} & KILE &  .415 (.021) & .349 (.019) & .280 (.015) & - & - \\
& KALE & {\bf .414} (.021) & {\bf .335} (.019) & .259 (.015) & - & - \\
& NNME\_GM4 & .455 (.027) & .364 (.024) & .258 (.018) & {\bf .167} (.019) & {\bf .108} (.009) \\
& NNME\_NICE &  .483 (.049) & .398 (.039) & {\bf .250} (.020) & .202 (.033) & .124 (.014) \\
\midrule  
\multirow{4}{*}{$\sigma_0=0.2$} & KILE & {\bf .773} (.040) & .711 (.038) & .682 (.042) & - & - \\
& KALE & .809 (.040) & .748 (.038) & .714 (.042) & - & -\\
& NNME\_GM4 & .820 (.081) & .696 (.066) & .548 (.037) & {\bf .389} (.024) & {\bf .313} (.027) \\
& NNME\_NICE &  .815(.065) & {\bf .670} (.058) & {\bf .459} (.050) & .429 (.045) & .350 (.035) \\
  \midrule
  \midrule
\multirow{4}{*}{$\sigma_0=0.1$} & KILE &  {\bf .226} (.020) & .187 (.019) & .150 (.009) & - & - \\
& KALE & .232 (.020) & .187 (.019) & .147 (.009) & - & -  \\
& NNME\_GM4 & .251(.027) & {\bf .169} (.019) & {\bf.116} (.009) & {\bf .079} (.005) & .076 (.009) \\
& NNME\_NICE &  .303 (.030) & .175 (.022) & .140 (.013) & .103 (.013) & {\bf .072} (.010) \\
\midrule 
\multirow{4}{*}{$\sigma_0=0.2$} & KILE & {\bf .376} (.037) & .362 (.037) & .351 (.034) & - & -  \\
& KALE & .382(.037) & .371( .037) & .363 (.034) & - & -\\ 
& NNME\_GM4 & .400 (.042) & {\bf .321}(.028) & {\bf.268} (.035) & {\bf .192} (.027) & {\bf.159} (.018) \\
& NNME\_NICE &  .442 (.042) & .367 (.052) & .300 (.055) & .241 (.039) & .181 (.021) \\
\bottomrule 
\end{tabular}}
\end{table}

The data generating process in the first setting is exactly the same as what KALE assumes, and one would expect KALE to perform the best. This is true when the measurement error is relatively small ($\sigma_0=0.1$) or the sample size is small to moderate ($n\in \{500,1000\}$). However, when the measurement error is large ($\sigma_0=0.2$) or the sample size is $n=2000$, NNME outperforms KALE. Additionally, between the two versions of NNME, NNME\_NICE has a better performance in this setting. We note that the two kriging methods, KILE and KALE, require computing the inverse of an $n\times n$ matrix. Therefore, they do not scale to large $n$. In fact, we are not able to run these methods for $n\in \{4000,8000\}$. In the second setting, $f$ is from the maximum of two Gaussian processes, and it is no longer the model assumed by the kriging methods. NNME uniformly outperforms KILE and KALE when $n>500$.  
Between the two versions of NNME, NNME\_GM4 is better in this setting.

We also investigated many similar settings with $f$ generated from 2-dimensional Gaussian processes, where we varied the RBF kernel and the distribution of measurement errors. A common observation is that the advantage of NNME, compared with kriging methods, is primarily in the cases where the sample size is large and  the measurement error is moderate to large. If the sample size is too small (e.g., $n\leq 500$) or the measurement error is too small,  NNME may have the over-fitting issue and tends to under-perform KILE and KALE (these kriging methods have fewer parameters to estimate and no over-fitting issues). In such cases, it is recommended to increase the L2-regulation parameters in NNME to counter  overfitting.

\subsection{Summary and discussions} \label{subsec:summary}
The numerical studies reveal several appealing properties of the neural network approach for estimating MEMs. The first is the flexibility of accommodating various kinds of $f(x)$.
%including smooth functions in classical nonparametric statistics, functions generated by Gaussian processes, and functions generated by neural networks. 
 This is because deep neural networks have the ability of representing a variety of function classes. In the simulations, we find that NNME has a reasonably good performance in almost all settings, provided that the decoder network has 7 or more layers. However, classical nonparametric/semi-parametric methods have unsatisfactory performance when $f(x)$ is non-smooth (Example 2).  %where $f(x)$ is generated by a Gaussian process with a mixture of two kernels.   

The second is its insensitivity to variation of tuning parameters and network structures. Classical methods require choosing the bandwidth or knots. Inappropriate choices will lead to oversmoothing or undersmoothing. Hence, tuning parameter selection crucially affects their performances. We tested a few data-driven tuning parameter selectors, such as the method in \cite{Delaigle2008} that uses SIMEX to select bandwidth in the local polynomial deconvolution method, but they worked unsatisfactorily. For the neural network approach, parameters to select include the L2 regularization parameters and the network structure (e.g., how many layers in the encoder/decoder and how many nodes per layer). The performance is relatively insensitive to these parameters. We often fix the L2 regularization parameter (e.g., $10^{-4}$ or $10^{-5}$) in simulations, and it works universally well. For the choice of the network structure, 
%what matters most is the number of layers in the decoder network; however, 
we find that the algorithm's performance is always similar as long as the decoder network has 7 or more layers. 

The third is the convenience of implementation for dimension $d>1$. In principle, one can extend most classical methods 
%for estimating measurement error models 
from $d=1$ to $d\geq 2$. However, several challenges remain: How to select the $d$-dimensional bandwidth matrix or the knots of $d$-dimensional splines? How to compute the solution? (Take the spline approach for example: The number of spline bases grows exponentially fast with $d$, posting a big challenge on computation.) Even for $d=2$ it is hard to implement most classical methods. In contrast, the neural network approach can easily handle $d>1$. 

The last, but not the least, is the scalability to large sample size $n$. For example, in Experiment 3, there are no other methods that can be  implemented for $n> 2000$, but the neural network approach can handle $n\in \{4000, 8000\}$ or even larger with ease. This is because the gradient ascent algorithm, and the back prorogation algorithm for computing gradients, is very fast. 
%{\color{red}(it is better to provide some numbers here, like how many seconds/minutes it takes for different methods)}

On the other hand, the neural network approach is not a panacea. Its main limitation is the lack of theoretical guarantee. The elegant theory for classical methods tell us when those methods are consistent and what their rates of convergence are \citep{Fan1993,Delaigle2009,Jiang2018}. For the neural network approach, so far we can only evaluate its performance numerically. Some limited theoretical understanding includes \cite{Kohler2011}, where they study a special least-squares neural network estimator and show that its rate of convergence with error-in-variables is similar to the rate of convergence in the error-free case, when the measurement error is small. However, their neural network estimator is a simple one that ignores measurement error, and their theory does not cover the case where measurement error is large, for which   NNME has the most appealing empirical performance.

\section{Real Data Applications} \label{sec:RealData}
We apply the NNME algorithm in two real data examples, the sea level study and the Framingham heart study.

\subsection{Sea level data}
The sea level studies model the change of sea levels in the past and predict the sea levels in the future. The data set in \cite{kemp2011climate} consists of measurements of relative sea level (RSL)  in North Carolina for the past 2000 years. The measurements were constructed from cores of coastal sediment, where ages of discrete depths in the core were estimated from radiocarbon dating and had uncertainty. Following \cite{Cahill2015}, we use a measurement error framework to account for age uncertainty. Let $y_i$ be the observed RSL, $x_i$ the true calendar year, and $w_i$ the estimated calendar year (by fitting the radiocarbon dates with an age-depth model). Suppose the ocean level is $g(x_i)$. The true RSL is defined as the difference between ocean level and land level. According to the glacio-isostatic adjustment, the land level decreases at an annual rate of $0.001\times r$, where $r$ is given.\footnote{The value of $r$ varies with geographical locations. This dataset was measured at two sites, where $r=0.9$ and $r=1$, respectively.} 
%This is called the glacio-isostatic adjustment (GIA). With GIA, the model becomes 
Then, the model is
\begin{equation} \label{eq:sea_model}
w_i = x_i + u_i, \qquad y_i = \underbrace{g(x_i) -[ c_0 + r\cdot (2.010 - x_i)]}_{\equiv f(x_i)} +\, \epsilon_i,
\end{equation}
where the calendar years $x_i$ and $w_i$ have been divided by 1000 (e.g., 1996 is written as 1.996) and $c_0$ is the land level in 2010 AD. Both $u_i$ and $\epsilon_i$ are assumed to follow normal distributions:
\begin{equation} \label{eq:sea_model_var}
u_i\sim {\cal N}\left(0, \sigma^2_{u_i}\right), \qquad \epsilon_i\sim {\cal N}\left(0, \tau^2 + \sigma^2_{\epsilon_i}\right), 
\end{equation}
where $\sigma^2_{u_i}$ and $\sigma^2_{\epsilon_i}$ are known for each observation. The data are $\{(w_i, y_i, \sigma^2_{u_i}, \sigma^2_{\epsilon_i})\}_{1\leq i\leq n}$. The goal is estimating the function $g(x)$. Without loss of generality, we let $c_0=0$, so that the ocean level in 2010 AD is viewed as the baseline. 

This problem is easily cast as a nonparametric regression with measurement errors. Write $f(x)=g(x)-r(2.010-x)$. We first apply NNME to estimate $f(x)$ and then convert it to an estimate of $g(x)$ straightforwardly. A minor difference from the previous MEMs is that the variance of response errors and measurement errors are both heterogeneous across observations. However, since $\sigma^2_{u_i}$ and $\sigma^2_{\epsilon_i}$ are known, our algorithm can be easily extended to this case. We incorporate $(\sigma^2_{u_i}, \sigma^2_{\epsilon_i}, \tau^2)$ into the marginal likelihood according to model~\eqref{eq:sea_model}-\eqref{eq:sea_model_var}, and modify the gradient ascent steps accordingly. These modifications are straightforward and omitted. We also modify \eqref{eqn:sy} to an estimate of $\tau^2$ as
\[
    \hat\tau^2 = \frac1{n}\sum_{i=1}^n\left\{\frac1{K}\sum_{k=1}^K (y_i - f(x_{ik}))^2 \frac{\beta_{ik}}{\sum_{\ell=1}^K \beta_{i\ell}} - \sigma^2_{\epsilon_i}\right\},
    %\label{eqn:sea_sy}
\]
where $x_{i1},x_{i2},\ldots,x_{iK}$ are importance samples for the $i$th observation and $\beta_{ik}$ is ratio between the complete likelihood and the density of proposal distribution, similar to that in \eqref{eqn:sy}. We use a parametric model
%working prior 
for $X$, %which is a shifted $t_2$ distribution with mean 0.75. 
which is a 2-component mixture of Gamma distributions with unknown parameters.
For the neural network structure, we let the decoder and encoder have 5 and 3 hidden layers, respectively, with 32 nodes per layer. We use tanh as the activation function in the decoder (to make $\hat{f}$ smooth) and ReLU in the encoder. Since the variances of measurement errors are small (from $2.5\times 10^{-7}$ to $0.009$) compared to the variance of calendar years (about $0.33$), we also add a direct link from $W$ to $\mu_{\phi}$ in Figure~\ref{fig:nn2} to accelerate the learning process. It means the output of encoder becomes $\mu(w, y) = \mu_\phi(w,y) + w$, where $\mu_\phi$ is from a 3-layer FNN. Besides an estimate of $f$, we also compute a 95\% confidence band via a parametric bootstrap procedure. In the standard model-based bootstrap, we should draw samples $\{(x_i^*,w_i^*, y_i^*)\}_{i=1}^n$ from the estimated model for $X$ and model \eqref{eq:sea_model}-\eqref{eq:sea_model_var} for $(W,Y)$. However, the noise variances in covariates and responses are only know at observed sites, and so we cannot use the standard model-based bootstrap. We instead fix $x_i^*=\mu(w_i,y_i)$, the estimated posterior mean of $x_i$ from the encoder, and draw $(w_i^*, y_i^*)$ from model \eqref{eq:sea_model}-\eqref{eq:sea_model_var} using the known variances for the $i$th observation. The resulting bootstrap confidence band can be viewed as conditioning on (estimates of ) $x_i$'s.

\begin{figure}[!hbt]
\centering
\includegraphics[width=.45\textwidth]{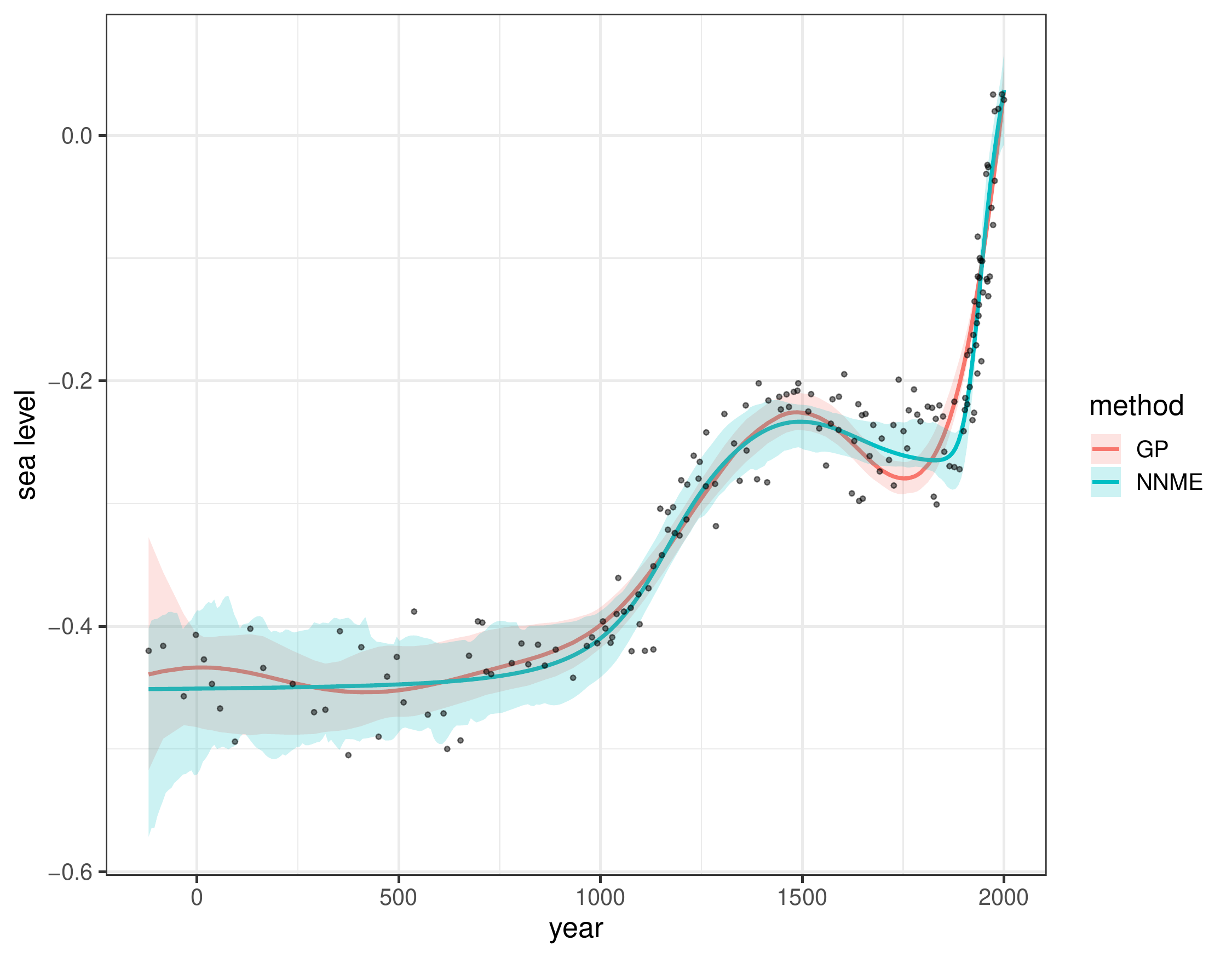}
\includegraphics[width=.45\textwidth]{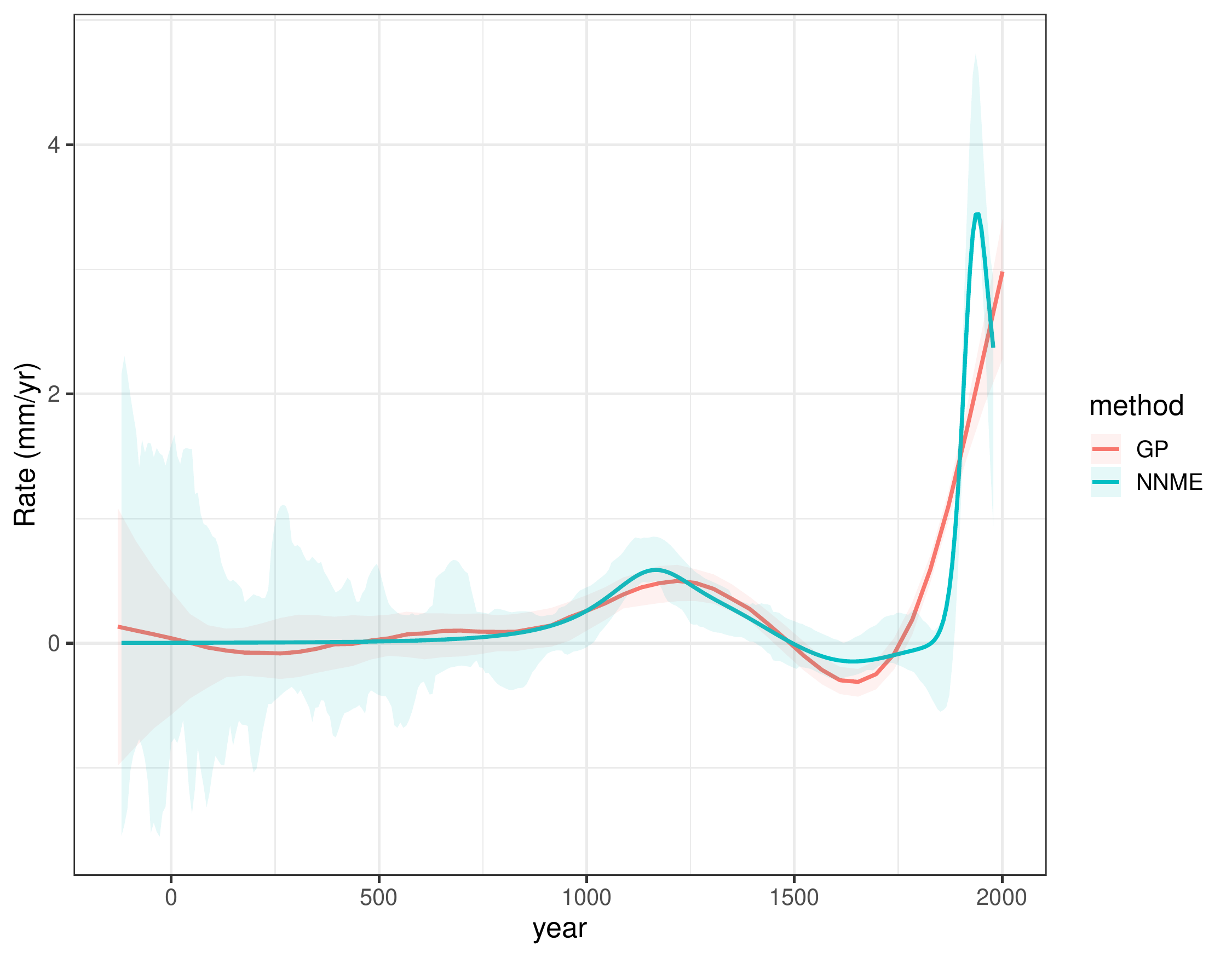}
\caption{Estimated sea level (left panel) and estimated rate of sea level change (right panel). The 95\% confidence band for NNME and 95\% credible interval for GP are shown.}\label{fig:sealevel2}
\end{figure}

The left panel of Figure~\ref{fig:sealevel2} shows the estimated ocean level, which is $\hat{g}(x)=\hat{f}(x)+r(2.010-x)$. The right panel shows the estimated sea level change, which is the derivative of $\hat{g}$. It is computed by a back propagation algorithm using estimated parameters. The confidence band of $\hat{g}$ is from a similar bootstrap procedure. We compare NNME with the approach in \cite{Cahill2015}, denoted as ``GP'', which models the sea level change (i.e.,  derivative of $g$) by a Gaussian process with measurement errors and uses Markov chain Monte Carlo for estimation and inference. While it is based on Gaussian process, this method is different from KALE.
The estimated curves of sea level (left panel of Figure~\ref{fig:sealevel2}) by NNME and GP are similar, except in the period between 1600 AD and 1800 AD. Both methods estimate the sea level to decrease first and increase later in this period, but NNME estimates this ``fluctuation'' be less prominent. For GP, we plot the 95\% Bayesian credible interval \citep{Cahill2015}. Although the credible interval is not directly comparable with the confidence band, we may still draw the conclusion from the plots that NNME gives less ``confidence'' on the fluctuation of sea level between 1600 AD and 1800 AD.    
The estimated curves of sea level change (right panel of Figure~\ref{fig:sealevel2}) are also similar. The curve by GP is smoother. A possible explanation is that GP directly models the sea level change while NNME models the sea level. 

Since there is no ground truth, we compare the performance of different methods using prediction errors. Given $\hat{f}$ and a testing $w$, we predict $y$ by $\mathbb{E}[\hat{f}(x)|w]$, where the expectation is with respect to the posterior distribution of $X$ given $W$. We use two ways to approximate the posterior distribution of $X$. The first is ${\cal N}(w,\sigma_0^2)$, where $\sigma_0^2$ is the variance of measurement errors in this observation. This ad-hoc approach can be viewed as imposing a flat prior on $X$. The second is using the estimated prior distribution of $X$ from NNME to derive the posterior distribution of $X$. In the actual   implementation, to obtain $\mathbb{E}[\hat{f}(x)|w]$, we first draw samples $\{x^*_i\}$ from the prior distribution of $X$ and then take a weighted average of $f(x^*_i)$, with weights  proportional to the conditional likelihood of $x_i^*$ given $w$. For each $\hat{f}$, we construct predictors using both ways. We measure the prediction performance using a cross-leave-out procedure: We randomly partition data into 5 folds, successively use each fold for testing (with the other 4 folds as training data) and compute the mean-squared-error. We then take the average of 5 mean-squared-errors and use it as the prediction error. We repeat this procedure for 10 times and report the mean and standard error of the prediction error. The results are as follows:

\begin{center}
\scalebox{.95}{
\begin{tabular}{ccccccccc} %|c|c|c|c|c|c|c|
 \toprule
NNME\_posterior1 & NNME\_posterior2 & GP\_posterior1 & GP\_posterior2 & NN \\ %& lpoly*  & Pspline (ideal tuning)\\
\hline
1.387 (0.068)  & 1.402 (0.069) &1.499 (0.065) & 1.503 (0.065) & 2.478 (0.020)\\  %& 1.29 (0.15) % & 0.94 (0.12)\\
%1.09 (0.07)  # if not weighted by p(x)
  \bottomrule 
\end{tabular}}
\end{center}
NNME is better than GP in terms of prediction performance. We also report the prediction error by NN; it ignores measurement errors in estimating $f$ and  predicts $f(x)$ by $\hat{f}(w)$. The performance of NN is much worse than NNME, suggesting that the advantage of NNME comes from not only using neural networks but also accounting for measurement errors.

\subsection{Framingham Heart Study}
The Framingham Heart Study is an ongoing cardiovascular study on residents of the town of Framingham, Massachusetts. The goal of this study is to predict whether the patient will have coronary heart disease (CHD) in 10 years. The dataset includes over 4,000 records and 15 attributes of patients including demographic, behavioral and medical risk factors.
We downloaded the data set from Kaggle\footnote{https://www.kaggle.com/dileep070/heart-disease-prediction-using-logistic-regression.}. 
The covariate of particular interest is the systolic blood pressure (SBP). However, it is impossible to measure the long-term SBP directly, and the observation recorded is the blood pressure measured in a single clinic visit, which has considerable daily variations. The measurement error of the transformed SBP (i.e., log(SBP$-50$)) was assumed to be Gaussian whose variance was estimated by several clinic visits of the same patient. Other covariates were assumed error-free, except for the logarithm of the total cholesterol level (Chol).
%which was assumed to have Gaussian measurement error  \citep{carroll2006measurement}. 
We followed \cite{carroll2006measurement} to model the measurement errors on SBP and Chol as bivariate Gaussian with a given covariance matrix. 
We proposed to fit a nonparametric logistic regression
\[
\mathbb{P}(Y=1) = L\Bigl(\beta_0+\sum_j\beta_jX_j+ f(X_{SBP}, X_{Chol}) \Bigr),
\]
where $L(x)$ is the logistic sigmoid function, $Y\in\{0,1\}$ indicates whether or not this patient has CHD in 10 years, $X_j$'s are error-free covariates, and for $(X_{SBP}, X_{Chol})$ only error-prone observations, $(W_{SBP}, W_{Chol})$, are available. Previous studies usually used a linear logistic regression model with measurement errors. In comparison, our approach can estimate the  nonlinear, interaction effects on SBP and Chol.

\begin{figure}[!hbt]
\hspace*{-1.5em}
\includegraphics[width=1.05\textwidth]{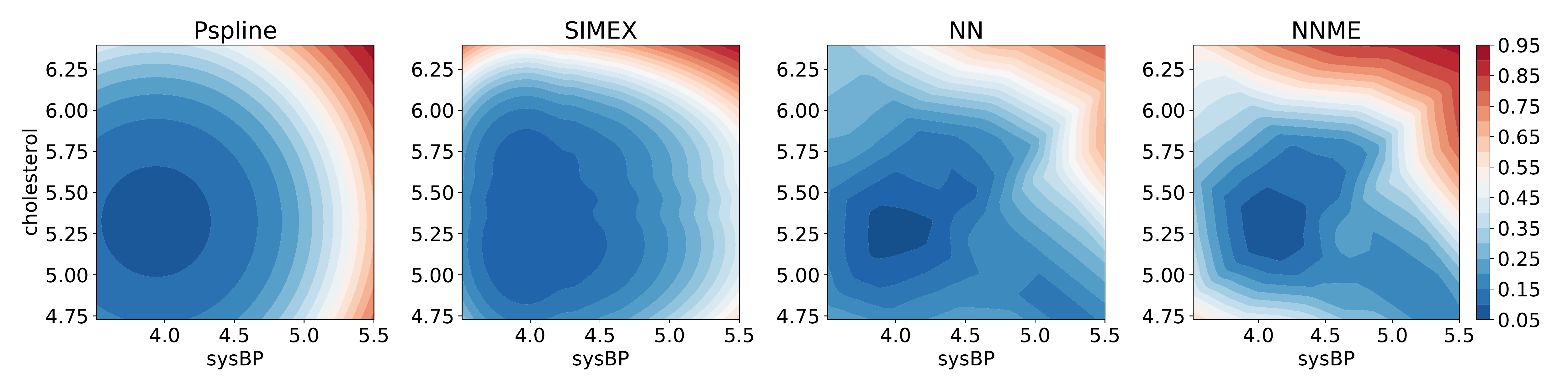}
\caption{Estimated 10-year risk of CHD as a function of SBP and Chol, where other covariates take the mean values. The contours are for a 2-dimensional function on $(x_{SBP}, x_{Chol})$, defined by $L\bigl(\hat{\beta}_0+\sum_j\hat{\beta}_j\bar{x}_j + \hat{f}(x_{SBP},x_{Chol})\bigr)$.}\label{fig:heart2}
\end{figure}

We pre-processed data by dividing the variable {\it age} by 100 and then centering all variables to have mean zero. We then adapted NNME to the current setting by changing the form of the log-likelihood. We estimated coefficients of other variables by jointly maximizing the IWAE objective via gradient ascent (again, we used the doublely reparametrized gradient estimator for parameters in the encoder).
We used 3 hidden layers for both the decoder and encoder, with 32 nodes per layer and ReLU as the activiation function.
To cope with the logistic model, we used the sigmoid function in the last layer of the decoder. We used NICE to model the (prior) joint distribution of SBP and Chol. The estimated 10-year risk of CHD, as a 2-dimensional function of SBP and Chol, is shown in Figure~\ref{fig:heart2}.

We compared NNME with the method SIMEX. SIMEX has to base on a nonparametric logistic regression method that ignores measurement errors; we used penalized splines (degree = 2, number of knots = 10).
\footnote{It is common to use quadratic extrapolation in SIMEX. However, on this data set,  quadratic extrapolation is considerably worse than linear extrapolation. Therefore, we reported results of linear extrapolation. }
The code to implement SIMEX basically allows for only 1 error-prone covariate; to apply it to 2 error-prone covariates, we have to assume an additive model. Consequently, this approach does not model interaction effects between SBP and Chol. In contrast, NNME includes a bivariate function of SBP and Chol in the risk and is able to capture interaction effects. 
Previously in Section~\ref{sec:Comparison}, we also considered KALE for 2 error-prone covariates; unfortunately, KALE has no direct extension to the logistic regression model. 
The estimated 10-year risks of CHD by NNME and SIMEX are shown in Figure~\ref{fig:heart2}. 
For a better comparison, we also considered two methods that ignore measurement errors, NN and Pspline, as counterparts of NNME and SIMEX. See Figure~\ref{fig:heart2}. The comparison of risk contours of Pspline versus NN suggests that neural neworks can capture nonlinear effects (beyond quadratic ones) of both SBP and Chol, as well as the interaction effect between two covariates. SIMEX generates more sophisticated risk contours than Pspline, but SIMEX still does not model interaction effects. NNME is the only one among 4 methods that accommodates measurement errors, nonlinear effects, and interaction effects.

To check whether NNME overfits,  
we evaluated the classification performance. Given an estimated model, we classify a sample by thresholding $\mathbb{E}[Y|\{X_j\}, W_{SBP}, W_{Chol}]$, where the posterior distribution of $(X_{SBP}, X_{Chol})$ is obtained from the measurement error distribution and the prior distribution from NICE. To avoid discussion of thresholds, we measure the performance by the area under ROC curve (AUC). 
We randomly selected 20\% of CHD samples and 20\% of non-CHD samples for testing and used the remaining samples for training. The mean and standard deviation of AUC, over 20 random splits of training and testing, are:
\begin{center}
\scalebox{.95}{
\begin{tabular}{cccccc} %|c|c|c|c|c|c|c|
 \toprule
 Pspline &   SIMEX & NN & NNME \\\hline %\_posterior2  &  NNME\_posterior1 
 0.727 (0.005)     & 0.728 (0.005) &     0.727 (0.005)    &  0.728 (0.005) \\ %&  0.726(0.007) \\
\bottomrule 
\end{tabular}}
\end{center}
The AUCs of different methods are similar. At least, it suggests that the more sophisticated models from neural networks are not due to overfitting. Additionally, accounting for measurement errors marginally improves the classification performance.

\section{Discussion} \label{sec:Discuss}

The use of neural networks in nonparametric statistics attracted a lot of recent attention, with encouraging progress on density estimation and nonparametric regression. This paper is an attempt to introduce neural networks to estimation of measurement error models, one of the classical topics in nonparametric statistics \citep{carroll1988optimal,Fan1993,Carroll1999}. 
We propose a neural network design, where the regression function $f(x)$, the (prior) distribution of $X$, and the posterior distribution of $X$ given the error-prone covariates are represented by three different neural networks. We estimate parameters of these neural networks by maximizing an ``importance sampling'' lower bound of the marginal log-likelihood of $(W,Y)$. We solve the optimization by a stochastic gradient ascent algorithm, with a doubly reparametrized gradient estimator. Our algorithm combines recent advancements in neural network, including \cite{Burda2015} on variational auto-encoder, \cite{Tucker2018} on stochastic gradient descent, and \cite{dinh2014nice} on normalizing flow.

Through extensive simulations and real data analysis, we demonstrated that the neural network approach is a promising alternative to classical nonparametric methods for MEMs. The neural network approach is flexible for accommodating various classes of functions (even non-smooth functions); its performance is insensitive to tuning parameters; it is convenient to implement for dimension $d>1$; and it has good scalability to a large sample size. Additionally, our method can be easily extended to more general settings. For example, if the noise in $X$ or the noise in $Y$ is non-additive, our method can be implemented similarly, where we simply change the expression of the joint density of $(X,W,Y)$. In contrast, classical nonparametric methods are more restrictive on model assumptions; for example, the deconvolution method \citep{Fan1993} relies on that the measurement error is additive. 

Theoretical understanding of neural network methods is a trending topic. Many theoretical frameworks were proposed, such as size-independent complexity \citep{Bartlett,golowich2018size},  implicit regularization \citep{soudry2018implicit}, seive approximation \citep{chen1999improved}, mean-field approximation \citep{mei2018mean}, and so on. Whether or not these theoretical frameworks can be used to understand the behavior of our method for MEMs is an open problem. We leave it for future work.

\vspace{0.4in}

\appendix
\begin{center} {\Large\bf Appendix}
\end{center}

\section{Variational inference versus maximizing likelihood}\label{app:other-methods}
\begin{figure}[!tb]
\centering
\includegraphics[width=0.7\textwidth]{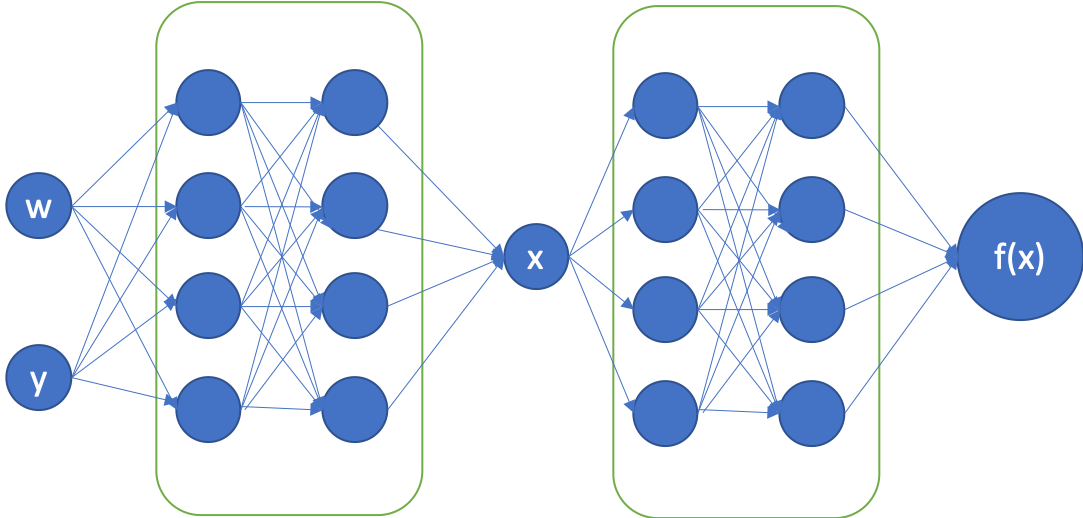}
\caption{The neural network structure for the maximizing joint likelihood framework. 
%The structure of Neural Network for Maximize joint likelihood method. In our simulations, the first green block consists of 3 fully connected layers (with nodes 32) with ReLu activation function, then the output is connected to linear layer to $x$. The second green block contains 4 fully connected layers (with nodes 32) with ReLu activation function, and the output is connected to a linear layer. 
}\label{fig:nn1}
\end{figure}

The neural network method we propose for MEMs adopts the variational inference framework, where parameters are estimated by maximizing an evidence lower bound (ELBO) of the marginal log-likelihood. In this appendix we consider a different approach, that is, maximizing the joint likelihood  of  $\{(x_i, w_i, y_i)\}_{i=1}^n$, with respect to both model parameters and unobserved values of $x_i$'s (MJL, henceforth). Same as before, we use an FNN to represent the regression function $f_\theta$. Next, we use another FNN to approximate the 
``most likely'' %{\it maximum a posteriori}  
estimate of $x_i$ given the observed data, i.e., 
\begin{equation}\label{NN1}
\hat{x}_i(\phi; w_i, y_i) = g_{\phi}(w_i, y_i), \qquad 1\leq i\leq n, 
\end{equation}
where $\phi$ contains parameters of this FNN, which will be determined jointly by $\{(w_j,y_j)\}_{j=1}^n$. We then write down the joint likelihood as (following the convention, we only present it for $n=1$; the extension to a general $n$ is straightforward): 
\begin{equation}\label{NN2}
\widetilde{L}(\theta,\phi|w,y ) =p_U\bigl(w - g_\phi(w,y)\bigr)\cdot p_{\epsilon}\bigl( y - f_\theta(g_\phi(w,y));\, \sigma^2  \bigr). 
\end{equation}
The parameters $(\theta,\phi)$ are estimated by maximizing $\widetilde{L}(\theta,\phi|w,y)$. The neural network structure that implements this estimator is shown in Figure~\ref{fig:nn1}.

In comparison, the variational inference framework approximates the posterior distribution of $x$ by ${\cal N}(\mu_\phi(w,y), \Sigma_\phi(w,y))$ and maximizes an integral with respect to this distribution, which serves as a lower bound of the marginal log-likelihood. See equation \eqref{eqn:obj}. 

We compare the ``maximizing joint likelihood'' framework and the variational inference framework. For a fair comparison, we use the basic variational inference approach, where the number of importance samples is $K=1$. It corresponds to the method VAE in Table~\ref{tb:NN-algs}. We solve both the MJL and VAE by the gradient ascent algorithm, where the gradient of \eqref{NN2} is computed directly, the gradient for VAE is estimated as in \eqref{eqn:plainGD} with $K=1$, and the step size for both is chosen by Adam \citep{kingma2014adam}. The noise variance $\sigma^2$ is optimized together with $(\theta,\phi)$, similarly as in Section~\ref{subsec:NNME}.

\begin{figure}[!htb]
\centering
\includegraphics[width=.45\textwidth]{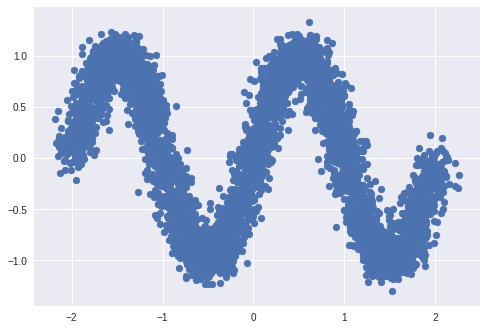}
\includegraphics[width=.45\textwidth]{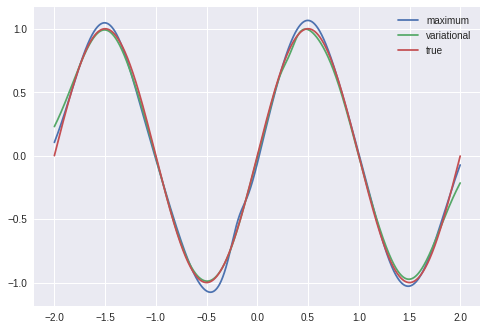}
\caption{Comparison of maximizing jointly likelihood (MJL) and variational inference (VAE). Settings are the same as in Experiment 1 of Section~\ref{subsec:otherversion}. Left: data. Right: fitted curves.}\label{fig:appendix-ML-VAE}
\end{figure}

We consider the simulation setting of Experiment 1 of Section~\ref{subsec:otherversion}, where $f(x)=\sin(\pi x)$ is the true regression function. The errors of MJL and VAE were already reported in Figure~\ref{fig:simu1_mise}(a). These results suggest that the variational inference framework is better than the ``maximizing joint likelihood'' framework,  especially when $n$ is small or moderate. 
%Here we investigate why the variational inference framework has an advantage. 
Intuitively the MJL framework ``imputes'' the ``missing data" to maximize the joint likelihood, whereas the VAE approaches marginalizing the ``missing data'' like in the EM algorithm, but based on a slightly incorrect distribution.
In Figure~\ref{fig:appendix-ML-VAE} we plot the estimated curves by MJL and VAE, where $n=5000$ and $\sigma_0=0.1$. At each $x$ where $f(x)$ reaches a local maximum or  minimum, the estimated curve by MJL has a bias in the neighborhood of $x$. This is because that MJL fails to take into account the  uncertainty of $X$ given $(W,Y)$. Intuitively, $\hat{f}(x)$ in MJL is determined only by those $(w_i,y_i)$ such that $g_\phi(w_i,y_i)\approx x$, while $\hat{f}(x)$ in VAE is determined by more $(w_i, y_i)$ corresponding to a wider range of $g_\phi(w_i,y_i)$. As a result, when $f(x)$ reaches a local minimum or maximum, it is likely to have extreme values of $y_i$, and even a few such $y_i$ can drag $\hat{f}(x)$  to be extreme in MJL; this will not happen in VAE because an individual value of $y_i$ is less influential.

\section{Comparison of models for $X$ in NNME} \label{subsec:LearnXdist}
The proposed method, NNME, is flexible to accommodate different kinds of models (prior distributions) for $X$. In Experiment 3 of Section~\ref{subsec:otherversion}, we tested 4 variants of NNME, where the model for $X$ is either the correct parametric model (2-component Gaussian mixture), or a misspecified parametric model (t-distribution, or 4-component Gaussian mixture), or a neural network model (NICE).

The data generation in this experiment is as follows: Fixing $\beta=16$, we generate $f$ from a 2-dimensional Gaussian processes as in \eqref{SGP}. In detail, we first generate $\{x_i\}_{1\leq i\leq n}$ from the distribution of $X$, which is a 2-component mixture of multivariate Gaussian distributions as in \eqref{simu-X-distribution}. Next, we construct an $n\times n$ covariance matrix $\Sigma$, with $\Sigma_{ij}=\exp(-\beta\|x_i-x_j\|^2)$, and then generate $(f(x_1),f(x_2),\ldots,f(x_n))$ from ${\cal N}(0, \Sigma)$. 

\begin{figure}[!tb]
    %\centering
    \includegraphics[width=1.05\textwidth]{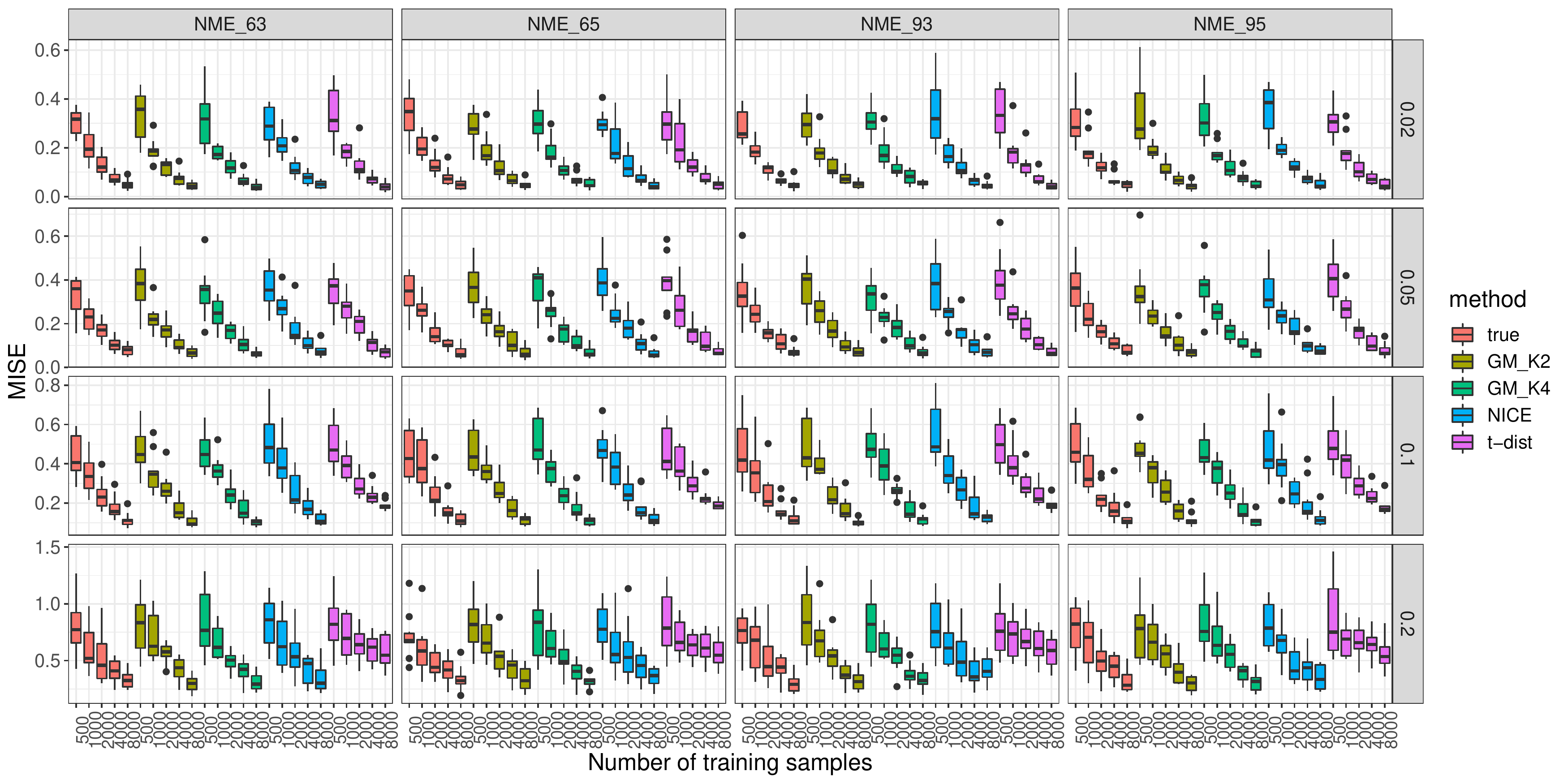}
    \caption{Comparison of different models for $X$ in NNME. $f$ generated from a 2-dimensional Gaussian process (see Experiment 3 of Section~\ref{subsec:otherversion}). Y-axis: ISE. X-axis: sample size $n$.}
    \label{fig:appendix-modelX}
\end{figure}

In Section~\ref{subsec:otherversion} we have reported the results for a few values of model parameters. Here we investigate more settings. We let the sample size $n$ range in $\{1000,2000,4000,8000\}$ and let the measurement error standard deviation range in $\{0.02, 0.05, 0.1, 0.2\}$. We also vary the neural network structures: Let $(\ell_1,\ell_2)$ be the number of layers of the two FNNs for representing $f_\theta$ and the posterior distribution of $X$, respectively. We let $(\ell_1,\ell_2)$ range in $\{(6,3), (6,5), (9,3), (9,5)\}$. The performance is measured by the integrated squared error (ISE) in the region $([-1, 0.2] \times [-1, 0.5])\cup([-0.5, 1] \times [-0.2, 1])$; according to the distribution of $X$ in \eqref{simu-X-distribution}, the probability of $x_i$'s falling outside this region is negligible, and so we restrict the error evaluation to be in this region. The results are shown in Figure~\ref{fig:appendix-modelX}, demonstrating that NNME performs robustly with different choices of the $X$ model when $n$ is small, but shows superiority with NICE or 4-component Gaussian mixture  when $n$ is large.

\section{Sensitivity to the depth of neural networks}  \label{app:CV}
As we stated in Section~\ref{sec:intro}, the neural network approach to MEMs is relatively insensitive to the choice of tuning parameters. Main tuning parameters include the depth of the two FNNs for representing $f$ and the posterior distribution of $X$. We now investigate the sensitivity of  NNME to the depth of neural networks. 

\begin{figure}[!t]
\centering
\includegraphics[width=.6\textwidth]{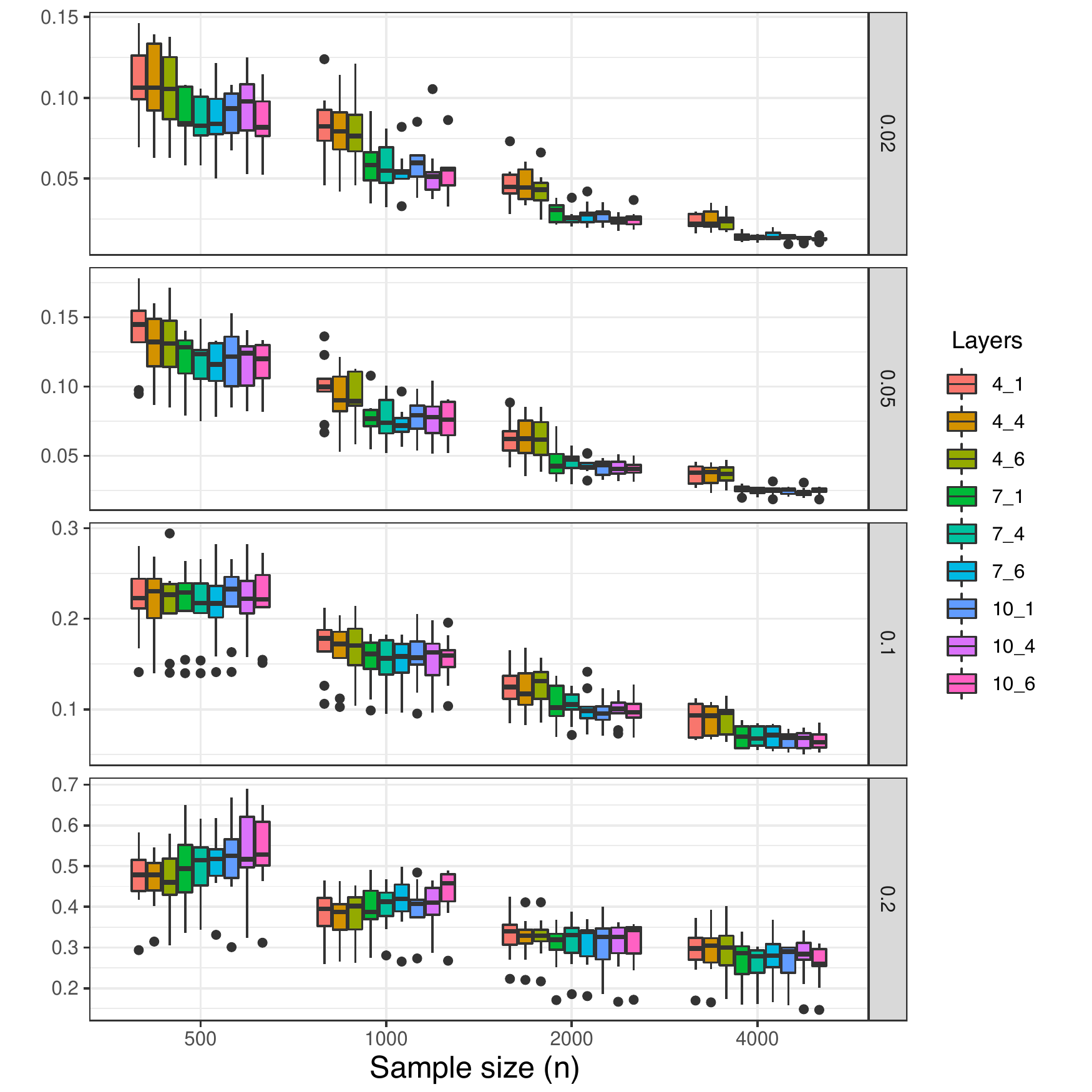}
\caption{Sensitivity to depth of neural networks. The name `7\_4' means that the decoder (for representing $f$) has 7 layers and the encoder (for approximating the posterior distribution of $X$) has 4 layers; other names are similar. Y-axis: ISE, x-axis: sample size. From top to bottom, $\sigma_0$ is $0.02$, $0.05$, $0.1$, and $0.2$, respectively.}  \label{fig:sensitivity}
\end{figure}

We consider a simulation setting where the $x_i$'s are drawn uniformly from $[-1,1]^2$ and the $f(x_i)$'s are generated from a Gaussian process as in \eqref{SGP} with $\beta=16$ (see Section~\ref{subsec:Simu3} or Appendix~\ref{subsec:LearnXdist} for details about generating data from a Gaussian process). The measurement errors are drawn from $N(0,\sigma_0^2)$. For a set of values of $(n, \sigma_0)$, we study the performance of NNME by varying the depth of neural networks. We place $\ell_1$ hidden layers in the decoder (for representing $f$) and $\ell_2$ hidden layers in the encoder (for approximating the posterior distribution of $X$), where each layer has 32 nodes with ReLU activation functions; the encoder has an additional layer with linear activation functions. We let $\ell_1$ range in $\{3,6,9\}$ and $\ell_2$ range in $\{0,3,5\}$. The integrated squared error (ISE) is evaluate using a $72\times 72$ grid.

The results are shown in Figure~\ref{fig:sensitivity}. For most values of $(n,\sigma_0)$, as long as the depth of neural networks satisfies $\ell_1\geq 6$ and $\ell_2\geq 3$, the performance is reasonably good. The only exceptions are when the sample size is small (e.g., $n=500$) or when the measurement error is large (e.g., $\sigma_0=0.2$). In such cases, we will see that a procedure like cross validation can select the appropriate number of layers.

\begin{figure}[!t]
\centering
\begin{subfigure}{0.48\textwidth}
\includegraphics[width=\textwidth]{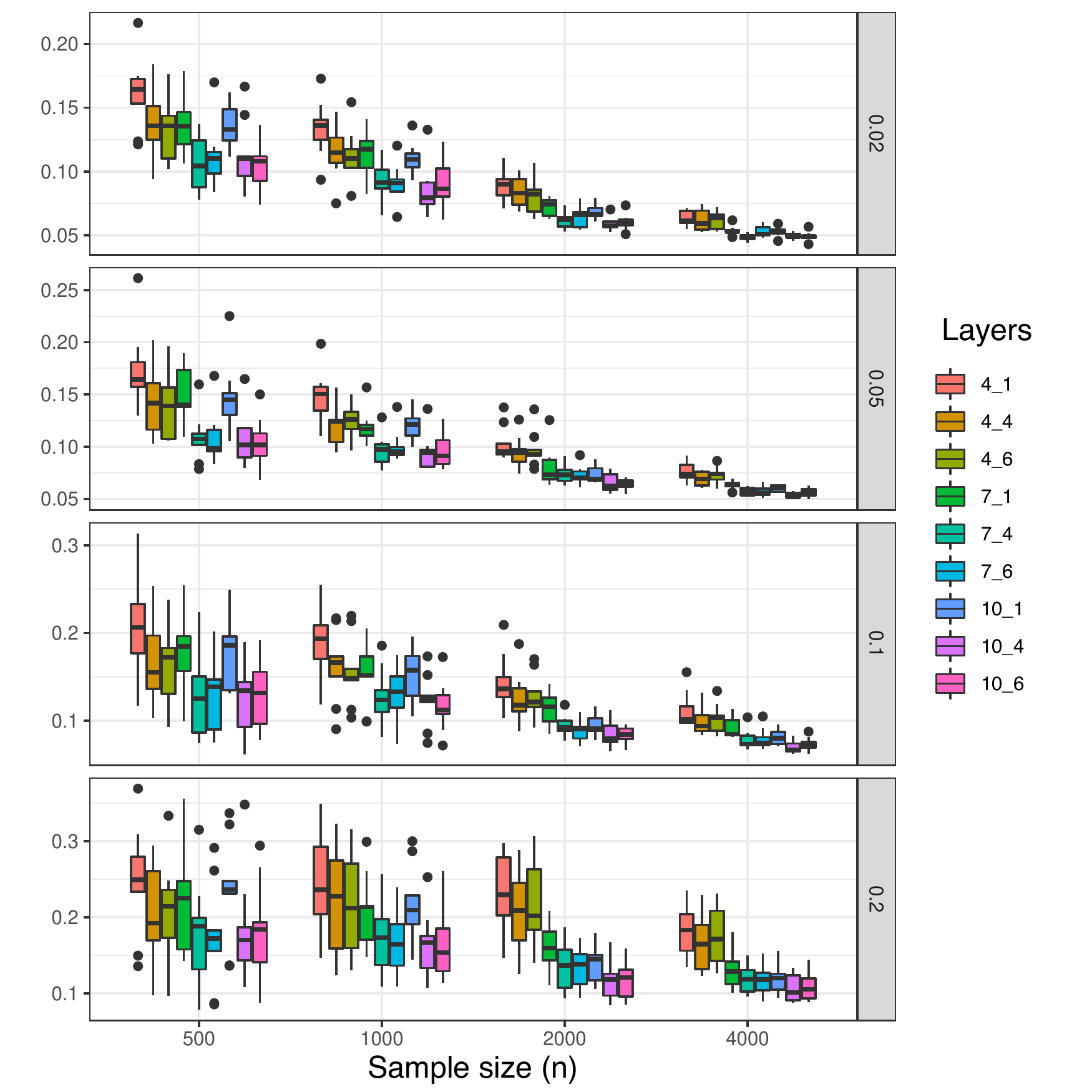}
\caption{Estimated RSS on validation data.}
\label{fig:test_err_val_var}
\end{subfigure}
\begin{subfigure}{0.48\textwidth}
\includegraphics[width=\textwidth]{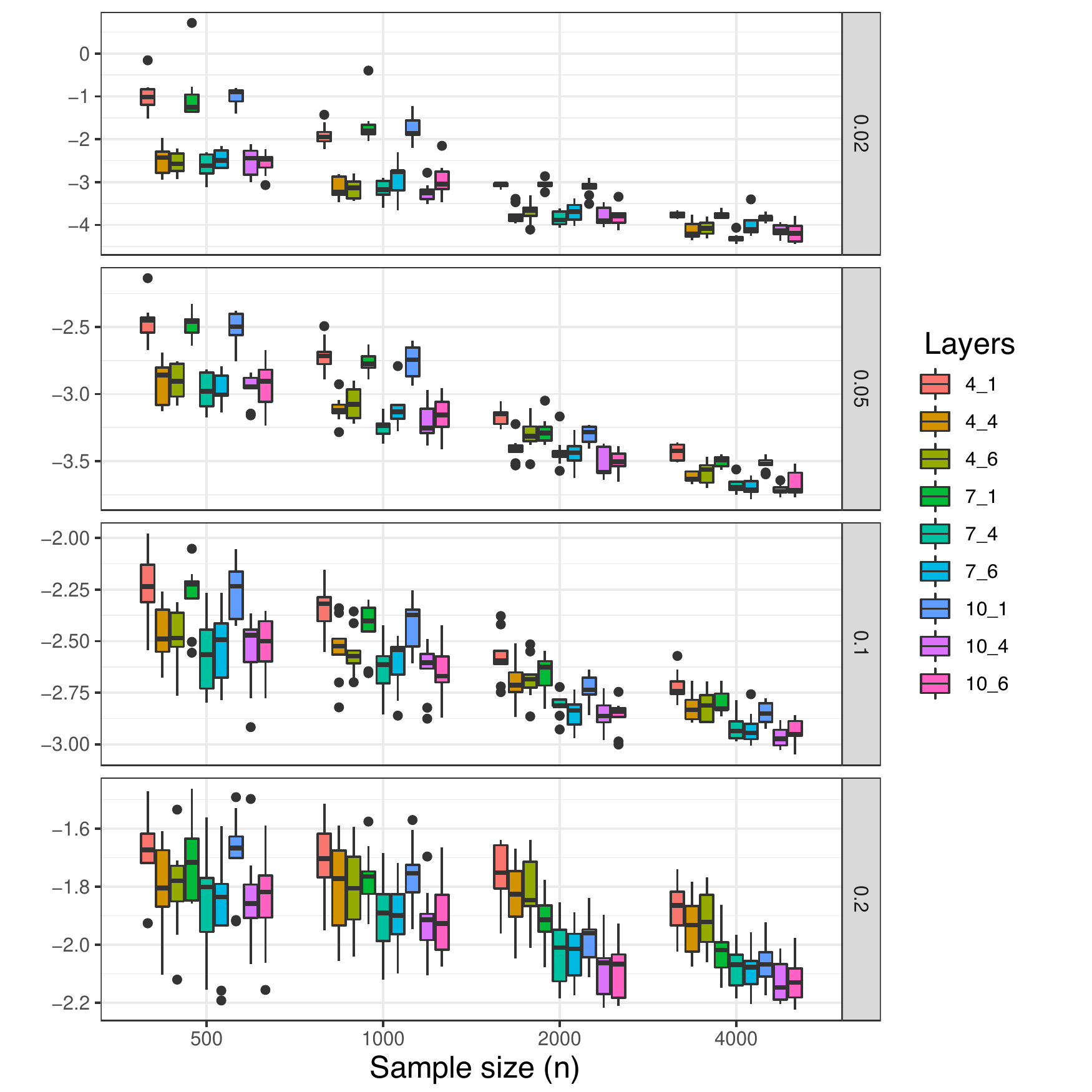}
\caption{Estimated ELBO on validation data.}
\label{fig:test_err_loss}
\end{subfigure}
\caption{Selection of depth of neural networks by different criteria. 
}\label{fig:CV}
\end{figure}

We then investigate the selection of depth of neural networks by two criteria. The first is the validation loss we eventually used, which is the weighted residual sum of squares in \eqref{eqn:sy}, evaluated on the validation data. To compute this quantity, we need to use estimated parameters from the training data and draw new Monte Carlo samples $z_{1:K}$; see Section~\ref{subsec:NNME} for details. We call this criterion the ``estimated RSS.'' 
%The second is the same loss evaluated on the training data. We call it the ``training RSS.'' 
The second is the loss function \eqref{eqn:obj} in NNME algorithm, evaluated on the validation data. It is an evidence lower bound (ELBO) of the marginal log-likelihood. Again, to obtain an approximation to this quantity, we need to use the fitted parameters from the training data and also draw new Monte Carlo samples $z_{1:K}$. We call it the ``estimated ELBO.'' For both criteria, we calculate the 5-fold cross-validation version. Figure~\ref{fig:CV} shows the values for different choices of depth of neural networks.

We compare these criteria with the true ISE in Figure~\ref{fig:sensitivity}. As we change the depth of neural networks, the trend in the estimated RSS roughly matches with the trend in the true ISE. Especially, as we mentioned above, if the sample size is small (e.g., $n=500$) or the measurement error is large (e.g., $\sigma_0=0.2$ and $n\leq 1000$), it is not always good to increase the depth of neural networks. In these cases, the cross-validation procedure, with the estimated RSS as the validation loss, can successfully guide us to the appropriate choice of depth.
The other criterion, the estimated ELBO, is a lot more sensitive to the parameters $\phi$ and tends to select a larger number of layers for the encoder. In comparison, the estimated RSS we used in Section~\ref{subsec:NNME} is a better option for the validation loss.

\bibliographystyle{apalike}
\bibliography{references}

\end{document}